\documentclass[runningheads]{llncs}
\usepackage{graphicx}
\usepackage{tikz}
\usepackage{comment}
\usepackage{amsmath,amssymb} % define this before the line numbering.
\usepackage{color}
\usepackage{booktabs}
\usepackage{multirow}

\usepackage{paralist}

\usepackage[ruled,vlined]{algorithm2e}

\usepackage[pagebackref,breaklinks,colorlinks]{hyperref}
\usepackage{breakcites}

\usepackage[accsupp]{axessibility}  % Improves PDF readability for those with disabilities.

\usepackage{wrapfig}

\makeatletter
\newcommand{\printfnsymbol}[1]{%
  \textsuperscript{\@fnsymbol{#1}}%
}
\makeatother

\begin{document}
% \linenumbers
\pagestyle{headings}
\mainmatter
\def\ECCVSubNumber{2327}  % Insert your submission number here

\title{ERA: Expert Retrieval and Assembly for Early Action Prediction} % Replace with your title

\begin{comment}
\titlerunning{ECCV-22 submission ID \ECCVSubNumber} 
\authorrunning{ECCV-22 submission ID \ECCVSubNumber} 
\author{Anonymous ECCV submission}
\institute{Paper ID \ECCVSubNumber}
\end{comment}
%******************

% \begin{comment}
% \titlerunning{ERA: Expert Retrieval and Assembly for Early Action Prediction}
\titlerunning{ERA for Early Action Prediction}
% \titlerunning{Expert Retrieval and Assembly}
\author{Lin Geng Foo\inst{1}\thanks{equal contribution} \and
Tianjiao Li\inst{1}\printfnsymbol{1} \and
Hossein	Rahmani\inst{2} \and
Qiuhong Ke\inst{3} \and
Jun Liu\inst{1}\thanks{corresponding author}
}
\authorrunning{L.G. Foo et al.}
\institute{ISTD Pillar, Singapore University of Technology and Design\\
\email{\{lingeng\_foo,tianjiao\_li\}@mymail.sutd.edu.sg,}
\email{jun\_liu@sutd.edu.sg}
\and
School of Computing and Communications, Lancaster University\\
\email{h.rahmani@lancaster.ac.uk}
\and
Department of Data Science \& AI, Monash University\\
\email{qiuhong.ke@monash.edu}
}
% \end{comment}
%******************
\maketitle

%%%%%%%%% ABSTRACT
\begin{abstract}
Early action prediction aims to successfully predict the class label of an action before it is completely performed. This is a challenging task because the beginning stages of different actions can be very similar, with only minor subtle differences for discrimination. 
In this paper, we propose a novel Expert Retrieval and Assembly (ERA) module that retrieves and assembles a set of experts most specialized at using discriminative subtle differences, to distinguish an input sample from other highly similar samples. 
To encourage our model to effectively use subtle differences for early action prediction, we push experts to discriminate exclusively between samples that are highly similar, forcing these experts to learn to use subtle differences that exist between those samples. 
Additionally, we design an effective Expert Learning Rate Optimization method that balances the experts' optimization and leads to better performance. We evaluate our ERA module on four public action datasets and achieve state-of-the-art performance.
% Code will be released.
\end{abstract}

\section{Introduction}

The goal of early action prediction is to infer an action category %the action's class 
at the early temporal stage, i.e., before the action is fully observed. This task is relevant to many practical applications, such as human-robot interaction \cite{reily2018skeleton,huang2016anticipatory,koppula2015anticipating}, security surveillance \cite{emad2021early,kong2018human,fatima2013unified} and self-driving vehicles \cite{gujjar2019classifying,mavrogiannis2020b,chaabane2020looking} since a timely response is crucial in these scenarios. 
% \lgg{To improve the utility of these applications (e.g enhanced safety on self-driving vehicles), prediction of actions before they are fully completed is necessary.} 
For example, for enhanced safety of self-driving vehicles, it is crucial that the actions of pedestrians can be predicted before they are fully completed, so that the vehicle can react promptly.
Such utility of early action prediction has not gone unnoticed, and it has received a lot of research attention recently \cite{li2020hardnet,tran2021progressive,hu2016real,wang2019progressive,weng2020early}.
% \lgg{(add more to the importance of the task)}
% \lgg{add even more citations. change the example}

\begin{wrapfigure}{r}{0.427\linewidth}
    \centering
    \includegraphics[width=\linewidth]{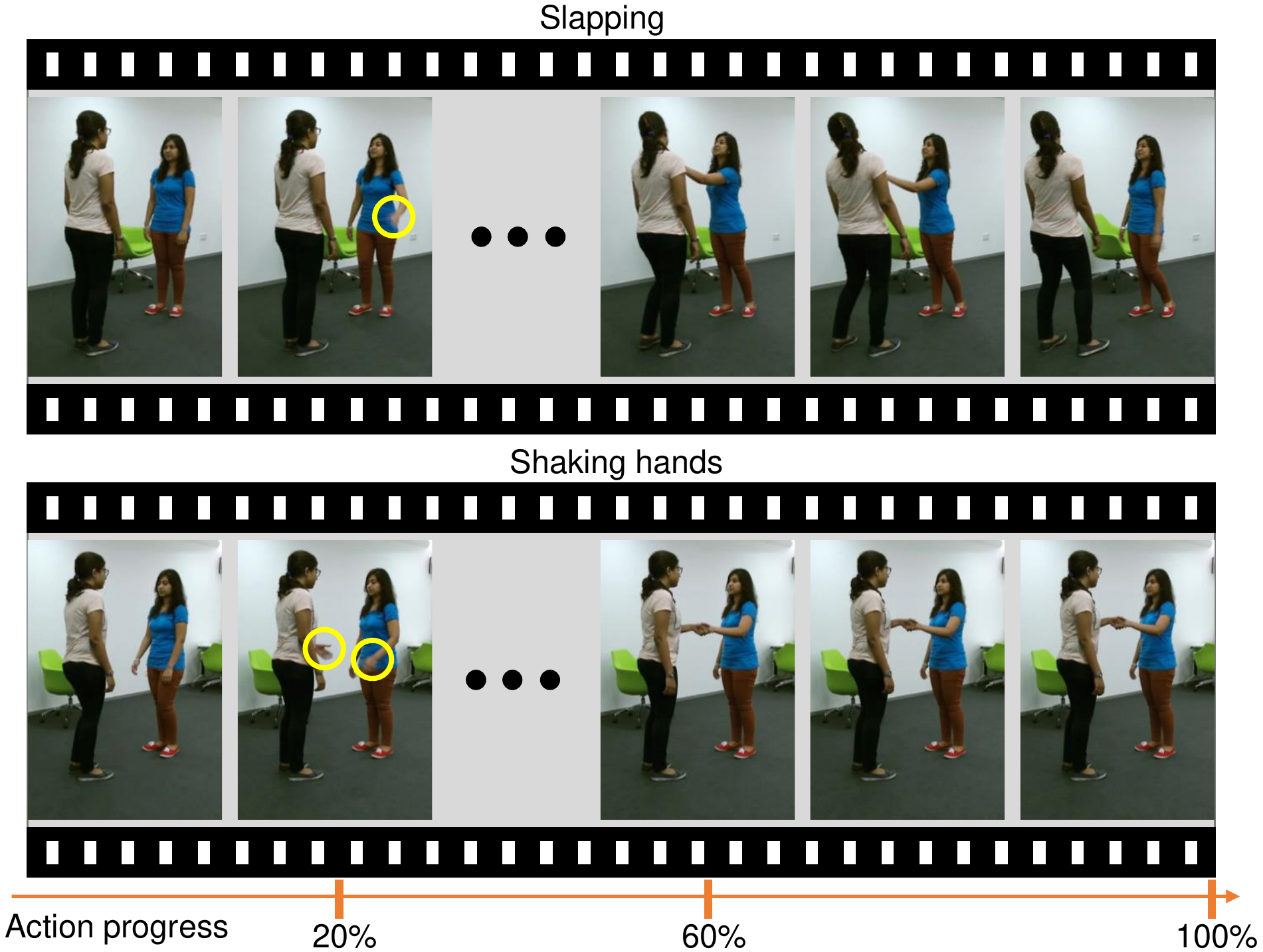} 
    % \vspace{-0.7cm}
    \caption{
    Illustration of two actions at the early stage, taken from NTU RGB+D 120 \cite{liu2019ntu}. Only subtle differences (highlighted in circles) exist for discrimination between the actions ``Slapping" and ``Shaking hands" at the early temporal stages (e.g., 20\%). 
    Best viewed in colour.
    % although these two actions are quite different when we look at the full sequence.
    }
    \label{fig:subtle_differences}
% \vspace{0.2cm}
\end{wrapfigure}

%% \cite{li2020hardnet,wang2019progressive,weng2020early}
Previous works \cite{li2020hardnet,weng2020early} show that one of the major challenges in early action prediction lies in the subtlety of the differences between some ``hard" samples at the very beginning temporal stages, since only limited initial observations of the action sequences are seen and some important discriminative information in the middle or later parts of the 
sequences is %are  % is
not observed, greatly increasing the difficulty of making correct predictions. %a correct prediction. 
For instance, as shown in Fig.~\ref{fig:subtle_differences}, though the human postures and motions in the full sequences of the actions ``slapping" and ``shaking hands" are quite different, their early parts are quite similar, with only subtle differences between them. 

%In Figure \ref{fig:subtle_differences}, we can observe that the two actions are easy to discriminate when the full sequences are observed, but are extremely similar at the beginning of the sequence, with only minor differences.

%left,bot,right,top
% \begin{figure}
% \end{figure}

To tackle early action prediction, various types of deep networks have been proposed  \cite{li2020hardnet,wang2019progressive,weng2020early}, but they still do not possess very good discrimination capabilities using subtle cues.
% In particular, deep neural networks prefer to learn to discriminate between the easier samples with major discriminative cues instead of the harder ones \cite{johnson2019survey}. 
In particular, deep networks prefer to learn to discriminate between the easier samples with major discriminative cues instead of the harder ones \cite{johnson2019survey}.
% , a tendency which we refer to as ``lazy" behaviour. %colloquially 
This can happen when we train the entire neural network by updating all its parameters using all samples -- the gradients update all the parameters to contribute towards correctly classifying all these samples, which thus can lead to the network learning more \textit{general patterns}  that apply to more samples, as opposed to learning \textit{specific subtle cues} to discriminate subtle differences that may only apply to a small subset of the data \cite{han2021dynamic}.
% Such performance drop has been noted for other applications where \textit{learning general patterns is insufficient}, 
% % such as segmentation of different instances of similar objects \cite{tian2020conditional}, 
% % such as instance segmentation of similar instances \cite{tian2020conditional}, 
% such as instance segmentation of multiple instances of the same object \cite{tian2020conditional}, 
% % or visual question answering of heterogeneous questions \cite{noh2016image}.
% or answering of heterogeneous visual questions \cite{noh2016image}.
% %
%
%
% This sub-optimal training behaviour of neural networks on subtle differences can be interpreted as some sort of ``lazy" behaviour by the neural network
The performance  drop
from such sub-optimal training behaviour can be further exacerbated on the very challenging action prediction task, where there can be a lack of major discriminative cues at the early stages among different actions, and the importance of utilizing subtle differences is increased.
% The performance drop from such ``lazy" behaviour can be further exacerbated on the very challenging action prediction task, where there is a lack of major discriminative cues at the early stages among different actions, and the importance of utilizing subtle differences is increased.
% \lgg{The performance drop from such sub-optimal training behaviour of neural networks can be further exacerbated on the very challenging action prediction task, where there is a lack of major discriminative cues at the early stages among different actions, and  the importance of utilizing subtle differences is increased. (seems strange, rephrase)}
% On the very challenging action prediction task, there is a lack of major discriminative cues at the early stages among different actions, and the importance of utilizing subtle differences (as opposed to general patterns) is increased, which also means that the performance drop from the usage of general patterns (originating from the sub-optimal training behaviour of neural networks) is high.
% The learning of general patterns is insufficient.
% Although recent work \cite{li2020hardnet} on early action prediction has attempted to improve the discriminative ability on subtle cues through mining hard samples to train the neural network, they train the parameters of the entire network using all samples, still leaving the network prone to sub-optimal performance with respect to subtle differences.
Although recent work \cite{li2020hardnet} on early action prediction has attempted to improve the discriminative ability on subtle cues through mining hard training samples, they train the parameters of the entire network using all samples, still leaving the network prone to sub-optimal performance with respect to subtle differences.
% \lgg{can cite some previous works on dynamic networks to further emphasize that our solution (versatility? using specialized set of convolutions) works. condconv, hydranet?}
%  question-aware parameters for visual question answering
%  \cite{noh2016image}
%  instance-aware parameters for instance segmentation
% image segmentation conditioned on different objects\cite{tian2020conditional}
% heterogeneous recognition task

% \vspace{-1em}
% \lgg{after removing the non-experts part. need to refine}
In this work, to improve the performance of deep networks on early action prediction, we propose an Expert Retrieval and Assembly (ERA) module that contains \textit{non-experts} and \textit{experts}.
% In this work, to improve the performance of deep networks on early action prediction, we propose an Expert Retrieval and Assembly (ERA) module that contains \textit{experts} across multiple Expert Banks.
Unlike \textit{non-experts} that contain parameters which are shared across all samples and capture general patterns that exist in many samples, \textit{experts} are only trained on a subset of the data (according to their \textit{keys}) and contain parameters that focus on encoding subtle differences to distinguish between highly similar samples.
% Unlike parameters which are shared across all samples and capture general patterns that exist in many easy samples, \textit{experts} are only trained on a subset of the data (according to their \textit{keys}) and contain parameters that focus on encoding subtle differences to distinguish between highly similar samples.
During the forward pass, a retrieval mechanism retrieves the most suitable experts, which are then assembled together with the non-experts to form a combination that is able to 
%, which is able to 
% \ke{making it capable to? (two which in the sentence)} 
discriminate samples using an effective mix of general patterns and subtle differences.
% During the forward pass, a retrieval mechanism retrieves the most suitable experts from the Expert Banks and assembles them together to form a combination, that is able to discriminate samples by encoding an effective mix of subtle differences.
% This retrieval mechanism is designed such that experts are retrieved by samples that are very similar, and thus, during training, the losses push the experts to learn \textit{specialized discriminative subtle cues to distinguish exclusively between these similar samples}, encouraging the acquiring of expertise in exploiting relevant subtle cues and alleviating ``lazy" behaviour.
This retrieval mechanism is designed such that experts are retrieved by samples that are very similar, and thus, during training, the losses push the experts to learn \textit{specialized discriminative subtle cues to distinguish exclusively between these similar samples}, encouraging the acquiring of expertise in exploiting relevant subtle cues.
% As a result, when we input a sample at test time, the retrieved experts will be highly specialized at exploiting relevant subtle cues for discrimination, which leads to improved performance on early action prediction.
The proposed ERA module is \textit{flexible}, and can be a plug-and-play replacement for traditional convolutional layers.

We design the ERA module with a set of experts to learn different subtle cues that exist across different actions.
However, it is non-trivial to balance the training among different experts in the ERA module, especially when the experts  
% each expert 
might be selected by vastly different numbers of samples. For instance, as some subtle cues may be more common, a few experts are selected more often and might be better trained. Such unbalanced training may limit the overall performance of our ERA module. A possible solution could be for each expert to have its own individual learning rate, 
and we adjust these learning rates such that the experts that require more training will have correspondingly higher learning rates. However, considering the numerous experts in the ERA module, coupled with the envisioned scenario where ERA modules replace multiple convolutional layers in a network, the number of hyperparameters is too large for manual tuning to be practical. Thus, we design an Expert Learning Rate Optimization (ELRO) method that balances the training of experts within the ERA module, improving the overall performance. 

% In summary, %our contributions are:
% the main contributions include:
In summary, our main contributions include:
% In summary, our contributions are:
\begin{inparadesc}
% \begin{itemize}
    \item[(1)] We propose a novel ERA module that effectively utilizes subtle discriminative differences between similar actions through retrieval and assembly of the most suitable experts for action prediction. 
    Our ERA module is a flexible plug-and-play module that can replace the traditional convolutional layer. 
    % \item[(2)] To balance the training among experts and further improve performance of the ERA module on early action prediction, we design an effective Expert Learning Rate Optimization method.
    \item[(2)] To balance the training among experts and further improve performance of the ERA module on early action prediction, we design an effective ELRO method.    
    % \item To minimize the need for tuning of Expert Bank sizes while achieving good performance and efficiency, we design our ERA module with a novel expansion capability.
    \item[(3)]  We obtain state-of-the-art-performance on early action prediction on four widely used datasets by replacing convolutional layers of the baseline architectures with our ERA modules.
% \end{itemize}
\end{inparadesc}

% \vspace{-0.1cm}
\section{Related Work}
% \vspace{-0.1cm}

\textbf{Early Action Prediction} refers to the task where only the front parts of each sequence are observed by the model. The loss of important discriminative information leads to a challenging scenario where subtle cues need to be properly utilized for successful discrimination. 
Different approaches \cite{li2020hardnet,tran2021progressive,hu2016real,wang2019progressive,ke2019learning,kong2018action,weng2020early,gammulle2019predicting,kong2014discriminative,kong2018adversarial,xu2019prediction,liu2019skeleton,kong2017deep,ke2017new,sadegh2017encouraging,pang2019dbdnet,chen2020recurrent,wang2021ga} have been proposed to address the early action prediction problem.
Li \textit{et al.} \cite{li2020hardnet} focused on the 3D early activity prediction task by mining hard instances and training the model to discriminate between them. 
Ke \textit{et al.} \cite{ke2019learning} proposed a Latent Global Network to learn how to transfer knowledge from full-length sequences to partially-observed sequences in an adversarial manner.
Weng \textit{et al.} \cite{weng2020early} introduced a policy-based reinforcement learning mechanism to generate binary masks to preclude the negative category information leading to improved recognition accuracy.
Wang \textit{et al.} \cite{wang2019progressive} proposed a teacher-student network architecture to distill the global information contained within the full action video
% \textcolor{red}{here we need a bit more detail. What do you mean by global information??? I assume you mean the video performing the full action. It needs to be clear}
from the teacher network to the student network.

In this work, 
unlike the above-mentioned methods, we explore the usage of a \textit{dynamic model} that pushes expert parameters to effectively encode subtle differences.
We propose a novel ERA module, which learns to discriminate among similar samples using subtle cues by retrieving and assembling relevant experts for each sample.

% we propose a novel ERA module, which learns to discriminate among similar samples using subtle cues by retrieving and assembling relevant experts for each sample.

\textbf{Action Recognition} is the task where a model predicts the classes of actions based on their full action sequences. Input data can come from different modalities, such as RGB data
% remove wang2016tsn and carreira2017i3d
\cite{lin2019tsm,feichtenhofer2019slowfast,xie2018s3d,yan2018participation,yan2020higcin,yan2020social} and skeletal data \cite{shi2019skeleton,shi2019two,chen2021ctrgcn,cheng2020skeleton,song2020stronger,ye2020dynamic,liu2020disentangling,shi2021adasgn,nguyen2021geomnet}. 
% \lgg{can cite more skeletal?}
Here, we focus on early action prediction, which is important yet more challenging since the early segments of different actions can be highly similar \cite{li2020hardnet,wang2019progressive,weng2020early}.
% Here, we focus on early action prediction, which is important yet more challenging because the early temporal segments of different actions can be highly similar \cite{li2020hardnet,wang2019progressive,weng2020early}.

% \textbf{Dynamic Networks} refer to neural networks that adapt their parameters or structures according to the input. A variety of different methods have been explored, including dynamic depth \cite{wang2018skipnet,veit2018convaig}, dynamic widths \cite{mullapudi2018hydranets,shazeer2017outrageously}, weight generation \cite{jia2016dynamic,tian2020conditional,noh2016image, yang2019condconv}, 
% % attention-aggregated kernels \cite{chen2020dynamic,yang2019condconv} 
% and spatially dynamic \cite{xie2020spatially,almahairi2016dynamic} methods. In general, dynamic networks can be employed for their improved computational efficiency and representation power.

\textbf{Dynamic Networks} refer to neural networks that adapt their parameters or structures according to the input. A variety of different methods have been explored, including dynamic depth \cite{wang2018skipnet,veit2018convaig,wu2018blockdrop}, dynamic widths \cite{mullapudi2018hydranets,shazeer2017outrageously}, weight generation \cite{chen2020dynamic, yang2019condconv}, 
dynamic routing \cite{wu2021coarse,li20212d}
% frame selection \cite{}
% model selection \cite{wu2021coarse}
% attention-aggregated kernels \cite{chen2020dynamic,yang2019condconv} 
and spatially dynamic \cite{xie2020spatially} methods. In general, dynamic networks can be employed for their improved computational efficiency and representation power.

As our ERA module retrieves a different set of experts for each input sample, it can be considered a type of dynamic module.
% Different from other methods, our ERA module focuses on tackling the \textit{early action prediction} task by pushing experts to \textit{discriminate exclusively between similar samples} to \textit{gain expertise in exploiting subtle differences}.
% Different from other methods \cite{chen2020dynamic,yang2019condconv,mullapudi2018hydranets,shazeer2017outrageously}, our ERA module focuses on tackling the \textit{early action prediction} task by pushing experts to \textit{discriminate exclusively between similar samples} to \textit{gain expertise in exploiting subtle differences}.
% To the best of our knowledge, our ERA module is the first work that dynamically retrieves experts to \textit{discriminate exclusively among subsets of similar samples during training}, 
% pushing them to \textit{gain expertise in exploiting subtle differences} to tackle early action prediction.
To the best of our knowledge, our ERA module is the first work that dynamically assigns experts to handle \textit{subsets of similar samples during training}, 
% tackling early action prediction by pushing them to \textit{gain expertise in exploiting subtle differences}.
% pushing them to \textit{gain expertise in exploiting subtle differences} to tackle early action prediction.
pushing them to \textit{gain expertise in exploiting subtle differences}.
% such that \textit{expertise is gained in exploiting subtle differences} to tackle early action prediction.
% To the best of our knowledge, our ERA module is the first dynamic module that pushes experts to be \textit{trained on only a subset of similar samples} that allows them to gain expertise in exploiting subtle differences.
This relies on our novel retrieval mechanism
involving key-query matching that retrieves experts to handle similar samples, which is different from existing mechanisms \cite{chen2020dynamic,yang2019condconv,mullapudi2018hydranets,shazeer2017outrageously}.
Moreover, we explore a novel ELRO method to further improve performance of our dynamic ERA module.
% \lgg{to edit. focus on subset}
% To the best of our knowledge, we are the first to train such experts, that only deal with a subset of similar samples, in order to push them to gain expertise.
% \lgg{Cite fine-grained paper?}
% \lgg{[no vspace? and need to delete more?]}

% \vspace{-0.1cm}
\section{Method}
% \vspace{-0.1cm}

Subtle differences among highly similar samples are difficult to be well-learned by deep neural networks that share all network parameters across all samples. 
When tackling the challenging early action prediction task, the importance of exploiting subtle cues is increased, as there can be a lack of major discriminative cues at the early stages of actions, which exacerbates the performance drop from the
sub-optimal performance of deep networks using subtle cues.
% sub-optimal training behaviour of deep networks on subtle cues.
% ``lazy" behaviour of deep neural networks.
% \ke{that learn major discriminative cues to classify  most easy samples?}.

% Motivated by this, we design a novel ERA module with an expert-retrieval mechanism to better exploit subtle cues. The expert-retrieval mechanism retrieves experts (from the Expert Banks) with relevant expertise for each input sample, and assembles them with the non-experts. 
% By matching experts with input samples that are highly similar to each other during training, this expert-retrieval mechanism allows experts to ignore distant samples, while pushing them to focus on distinguishing between highly similar samples by specializing in subtle differences.
Motivated by this, we design a novel ERA module with an expert-retrieval mechanism to better exploit subtle cues. The expert-retrieval mechanism retrieves experts (from the Expert Banks) with relevant expertise for each input sample, and assembles them with non-experts. 
By matching experts with input samples that are highly similar to each other during training, this mechanism allows experts to ignore distant samples, while pushing them to focus on distinguishing between highly similar samples by specializing in subtle differences.

Due to the uneven distribution of samples across different experts, there might be uneven training among experts, which limits performance of our ERA module.
% To mitigate this issue, an effective Expert Learning Rate Optimization method is implemented during the training of the experts, which tunes their individual learning rates, resulting in a more effective training of experts and improved performance. 
To mitigate this issue, an effective ELRO method is implemented during the training of the experts, which tunes their individual learning rates, resulting in a more effective training of experts and improved performance.

% In practice, we find that the sizes of Expert Banks are important hyperparameters which can significantly affect performance and efficiency. 
% Moreover, tuning of this hyperparameter for individual Expert Banks might be impractical when there are too many of them (there can be several within each ERA module, which can be placed throughout many layers in a network), and might each require different settings.
% To resolve this issue, we introduce our Expert Banks with an expansion capability, which automatically grows our Expert Banks to an optimal size during training, significantly reducing the need for hyperparameter tuning while maximizing performance and efficiency.

% Below, we first describe the early action prediction task. Then, we introduce the ERA module and explain in detail how the sample selection mechanism can encourage experts to specialize during training. Lastly, we describe our Expert Learning Rate Optimization method.
Below, we first describe the early action prediction task. Then, we introduce the ERA module and explain in detail how the expert-retrieval mechanism can encourage experts to specialize during training. Lastly, we describe our ELRO method.

% Below, we first describe the early action prediction task. Then, we introduce the ERA module and explain in detail how, during training, the sample selection mechanism can encourage experts to specialize while the Expert Bank expands automatically when new experts are beneficial for performance. 

\subsection{Problem Formulation}
% \textbf{Problem Formulation.}
A full-length action sequence can be represented as a set $S = \{s_t \}_{t=1}^{T}$ containing $T$ frames, where $s_t$ denotes the frame at the $t$-th time step. Following previous works \cite{hu2016real,ke2019learning,li2020hardnet}, $S$ is divided into $N$ independent segments, with each segment containing $\frac{T}{N}$ frames. A partial sequence consists of a set of frames $P = \{s_t \}_{t=1}^{\tau}$, with $\tau$ being the last frame in any one of the $N$ segments, i.e., $\tau = i \frac{T}{N}, i = \{1,2,...,N \}$. The task of early action prediction is to predict the class $c \in \{1,2,...,C \}$ of the activity that the partial sequence $P$ belongs to, and different observation ratios $\frac{\tau}{T}$ of $P$ are tested.

% \begin{wrapfigure}{r}{0.5\linewidth}
%     \centering
%     \includegraphics[width=\linewidth]{figs/cell_v20.pdf}
%     \caption{
%     % Illustration of the proposed ERA module. 
%     \lgg{change the diagram, to the new method. More expert banks}
%     Schema of our ERA module. 
%     % For clarity, only a single input $X$ is shown. 
%     In the Expert Bank, there are $M$ experts, with each expert $E_{ni}$ containing parameters $m_{ni}$ and a key $k_{ni}$.
%     % representing its area of expertise.
%     %and an assembly of $d$ retrieved experts.
%     The two important steps (i.e., retrieval and assembly) are indicated with red arrows. In the \textit{retrieval} step, a query $q$ generated from the mapping function $f$ is used to retrieve the most suitable experts from the Expert Bank through a key-query matching mechanism. 
%     In the \textit{assembly} step, the top $d$ retrieved experts are assembled and combined with the non-experts to produce the output $Y$.
%     }
%     \label{fig:deer}
%     % \vspace{-0.2cm}
% % \end{figure}
% \end{wrapfigure}
% % \vspace{-0.5em}
% \subsection{ERA Module}
% % \vspace{-0.5em}

% As shown in Fig.~\ref{fig:deer}, our ERA module consists of multiple Expert Banks of experts. 
% \lgg{change Expert Bank instances}

\subsection{ERA Module}
\begin{wrapfigure}{r}{0.565\linewidth}
    \centering
    \includegraphics[width=0.98\linewidth]{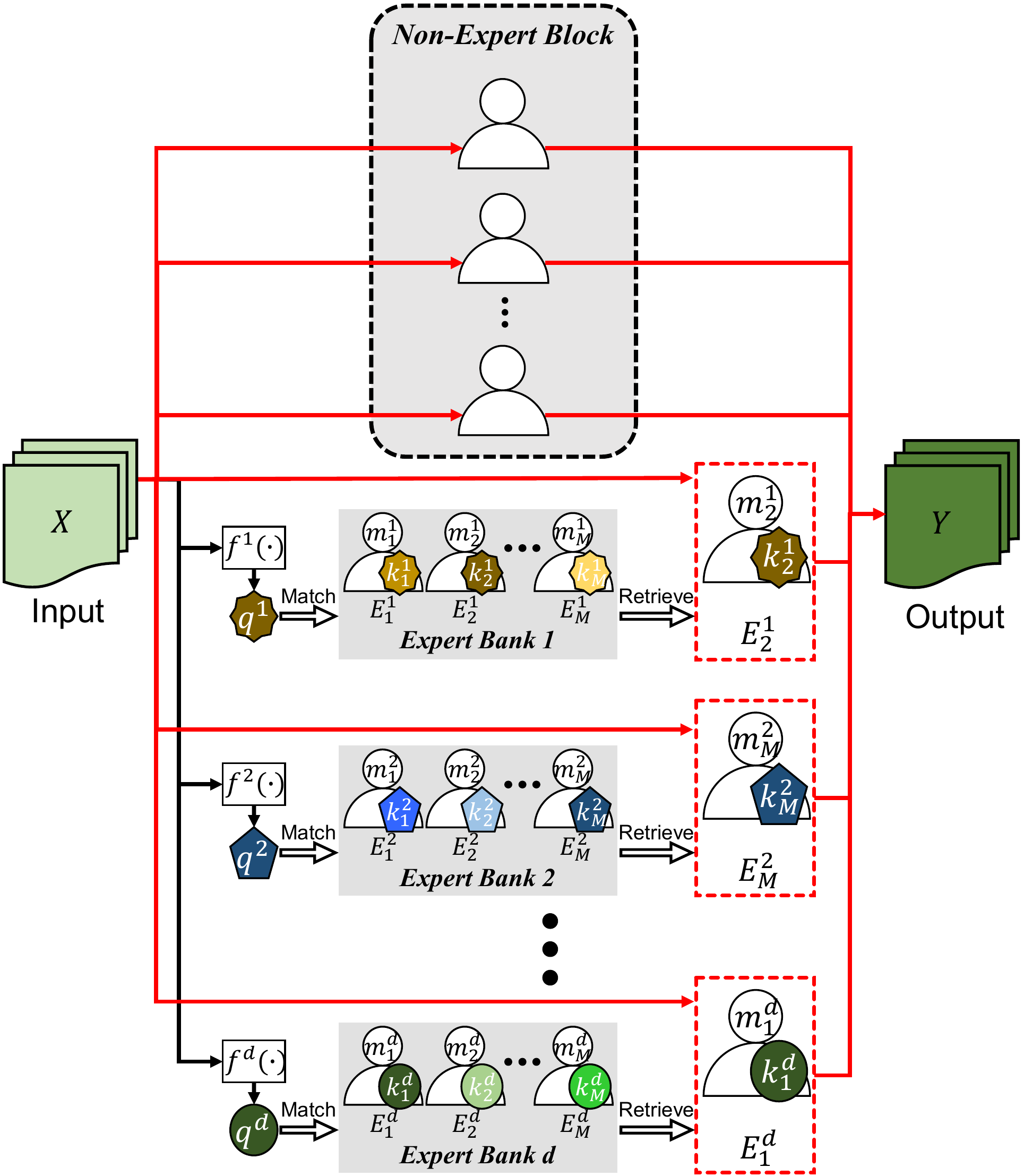}
    % \vspace{-0.5cm}
    \caption{
    Schema of our ERA module. 
    Our ERA module contains $N_{out}$ channels in total: $N_{out}-d$ non-expert channels (Top) and $d$ expert channels (Bottom).
    % For each expert channel, an expert is retrieved...
    Each expert channel retrieves its expert from a corresponding Expert Bank that contains $M$ candidate experts, where
    each expert $E_{i}^{p}$ consists of parameters $m_{i}^{p}$ and a key $k_{i}^{p}$.   
    The two important steps (i.e., retrieval and assembly) are indicated with red arrows.     
    In the \textit{retrieval} step, an expert will be retrieved from each Expert Bank through a key-query matching mechanism, such that $d$ experts are retrieved across $d$ expert channels.
    % queries (e.g., $q^{p}$) generated from mapping functions (e.g., $f^{p}$) are used to retrieve the most suitable expert from their corresponding (e.g., $p$-th) Expert Bank through a key-query matching mechanism.    
    In the \textit{assembly} step, the $d$ retrieved experts are assembled and combined with the $N_{out}-d$ non-expert kernels to produce the output $Y$ (with $N_{out}$ channels).    
    % Schema of our ERA module. 
    % Our ERA module contains $d$ Expert Banks with $M$ experts each.
    % Each expert $E_{i}^{p}$ contains parameters $m_{i}^{p}$ and a key $k_{i}^{p}$.   
    % % Each of the $d$ expert channels has a corresponding Expert Bank.
    % The two important steps (i.e., retrieval and assembly) are indicated with red arrows.     
    % In the \textit{retrieval} step, queries (e.g., $q^{p}$) generated from mapping functions (e.g., $f^{p}$) are used to retrieve the most suitable expert from their corresponding (e.g., $p$-th) Expert Bank through a key-query matching mechanism.    
    % In the \textit{assembly} step, the $d$ retrieved experts are assembled and combined with the non-experts to produce the output $Y$.    
    }
    \label{fig:deer}
    % \vspace{-0.5cm}  %for cosmetic purpose only, as it cuts into the next page
\end{wrapfigure}

% As shown in Fig.~\ref{fig:deer}, our ERA module consists of multiple Expert Banks of experts, an expert block and a non-expert block.
% As shown in Fig.~\ref{fig:deer}, our ERA module consists of an expert block, a non-expert block and candidate experts contained within multiple Expert Banks.
As shown in Fig.~\ref{fig:deer}, our ERA module consists of \textit{candidate experts} contained within multiple Expert Banks and a \textit{non-expert block}.
Considering that convolutional architectures have been shown to be effective for the early action prediction task \cite{li2020hardnet,wang2019progressive}, the experts are implemented as convolutional kernels.
For ease of notation, we describe our method in a 2D convolutional kernel setting, even though it can be generalized to 1D, 3D or graph convolutions as well. 
This ability to generalize to other types of convolutions is important, as existing early action prediction architectures often use various types of convolutions, such as 3D convolutions \cite{shi2019two} or graph + 2D convolutions \cite{hara2018can}.

Let an input be $X \in \mathbb{R}^{N_{in} \times N_h \times N_w}$, where $N_{in}, N_h$ and $N_w$ represent the channel, height and width dimensions of the input feature map. 
Note that here we omit the batch dimension for simplicity.
% Assume that the convolutional filter selected (to be replaced with our ERA module) has shape $N_{out} \times N_{in} \times b_h \times b_w$, where $N_{out}$ represents the number of output channels, and $b_h$ and $b_w$ represent the height and width of the convolutional kernel.
% Assume that, in the backbone model, input $X$ is processed by a convolutional filter of shape $N_{out} \times N_{in} \times b_h \times b_w$, where $N_{out}$ represents the number of output channels, and $b_h$ and $b_w$ represent the height and width of the convolutional kernel.
Assume that, in the backbone model, input $X$ is processed by a convolutional filter $W_{conv} \in \mathbb{R}^{N_{out} \times N_{in} \times b_h \times b_w}$, where $N_{out}$ represents the number of output channels, and $b_h$ and $b_w$ represent the height and width of the convolutional kernel.
% For ease of understanding, it might be helpful to see the convolutional filter as $N_{out}$ kernels of shape $N_{in} \times b_h \times b_w$ that respectively produce each of the $N_{out}$ output channels.
% \lgg{Here, we propose to}
% We aim to replace this convolutional filter with our ERA module, for better performance on early action prediction.
We aim to replace $W_{conv}$ with our ERA module, for better performance on early action prediction.

% Specifically, we design our ERA module to also ultimately produce weights $W_{ERA}$ of the same shape ($N_{out} \times N_{in} \times b_h \times b_w$) as that of $W_{conv}$, which can be seen as being $N_{out}$ kernels (each of shape $N_{in} \times b_h \times b_w$) that respectively produce each of the $N_{out}$ output channels.
Specifically, we design our ERA module to also ultimately produce weights $W_{ERA}$ of the same shape ($N_{out} \times N_{in} \times b_h \times b_w$) as $W_{conv}$, which can be seen as $N_{out}$ kernels (each of shape $N_{in} \times b_h \times b_w$) that respectively produce each of the $N_{out}$ output channels.
% More specifically, in our ERA Module, we split the $N_{out}$ kernels into two parts: $d$ expert kernels and $N_{out} - d$ non-expert kernels, where $d$ is a hyperparameter.
More specifically, in our ERA Module, we split the $N_{out}$ channels (and therefore also kernels) into two parts: $d$ expert channels and $N_{out} - d$ non-expert channels, where $d$ is a hyperparameter.
To allow our $d$ expert channels to specialize in subtle cues, we would like \textit{each expert to be trained on only a subset of the data}, thus we introduce $d$ Expert Banks containing $M$ candidate experts each, and retrieve only one expert from each Expert Bank per sample, such that the other $M-1$ candidate experts in the bank are unused for this sample.
The $N_{out} - d$ non-expert kernels (collectively defined in a non-expert block $W_{nonexpert} \in \mathbb{R}^{(N_{out} -d) \times N_{in} \times b_h \times b_w}$) are shared over all samples, and thus tend to learn general patterns. 
% We utilize a combination of both non-expert and expert parameters, because the usage of non-expert parameters to capture general patterns that exist in distant samples, is complementary with our experts that specialize at
% capturing subtle cues for discriminating between similar samples, and their combination leads to improvements on early action prediction.
We utilize a combination of both non-expert and expert kernels, because the usage of non-expert kernels to capture general patterns, is complementary with our experts that specialize at
capturing subtle cues for discriminating between similar samples, and their combination leads to improvements on early action prediction.

%550
% \vspace{-0.4cm}

%emphasize each channel. focus on one (for clarity)
%for each channel, we find the best expert with the most suitable expertise
%idea is to abstract each channel... not really the kernel
%combine these channels together

% \subsubsection{1) Expert Banks}
\noindent\textbf{1) Expert Banks} 
% \noindent\textbf{Expert Banks.} 
% \lgg{going to edit to multiple expert banks again, focus on multiple semantics between channels.}
% \lgg{Now editing to the new method, where a set of d experts are always selected together}
% \lgg{To change, to a 2D bank format.}
% \lgg{To edit to channel expansion method + growing}
% \lgg{To edit to multiple expert banks. different channels}
% \lgg{will rewrite this part to avoid the angle of multiple channels. maybe just focus on one expert bank and channel for clarity}
% \lgg{change this part. edit the notation to add superscripts. make things clear. add more motivations.}
% We define $d$ Expert Banks in each ERA module, each containing $M$ experts. 
% For each of the $d$ expert channels in the ERA module, we define an Expert Bank containing $M$ candidate experts.
% We define $d$ Expert Banks in each ERA module, each containing $M$ experts. 
% As shown in Fig.~\ref{fig:deer}, each Expert Bank contains candidate experts for a corresponding channel of the expert block, and these experts learn to \textit{encode subtle differences relevant to the semantics of that particular channel}.
% The $M$ candidate experts within the Expert Bank are trained on subsets of similar samples, allowing them to specialize in encoding \textit{subtle differences} among a subset of similar samples. 
% learn to encode subtle differences for different subsets of samples.
%
To facilitate our intention to let each expert be trained on only a subset of samples, we define $d$ Expert Banks, each containing $M$ candidate experts, as shown in Fig.~\ref{fig:deer}. %\textcolor{green}{I don't think this is correct "Each Expert Bank corresponds to an expert kernel".}
% Each Expert Bank corresponds to an expert kernel, and the candidate experts in each Expert Bank can be retrieved for that corresponding expert kernel. 
% Among the $M$ candidate experts in the $p$-th Expert Bank, one will be retrieved for the corresponding $p$-th expert kernel.
% The $M$ candidate experts in the $p$-th Expert Bank are all potential candidates that can be retrieved for the corresponding $p$-th expert kernel.
The $M$ candidate experts in the $p$-th Expert Bank are all potential candidates that can be retrieved for the corresponding $p$-th expert channel.
We define the $i$-th expert in the $p$-th Expert Bank as $E_{i}^{p}$, where $E_{i}^{p}$ contains convolutional kernel weights $m_{i}^{p}$ and a key $k_{i}^{p}$. The key $k_{i}^{p}$ is used for matching with the most suitable samples, and represents the \textit{area of expertise} of this expert, as it determines the samples that the expert will be retrieved for.
Meanwhile, the expert kernel $m_{i}^{p}$ acts as a \textit{specialized mechanism} to process the discriminative subtle cues on the input features that match the key $k_{i}^{p}$.
The expert key $k_{i}^{p}$ and kernel $m_{i}^{p}$ are model parameters that are trained in an end-to-end manner.
For each expert $E_{i}^{p}$, $m_{i}^{p} \in \mathbb{R}^{N_{in} \times b_h \times b_w}$ and $k_{i}^{p} \in \mathbb{R}^{K}$, where $K$ represents the dimensionality of the key, and $K << N_{in} \times N_h \times N_w$ for efficiency.

% \noindent\textbf{Expert Retrieval.} To retrieve experts, we first extract a compact and meaningful representation from the input feature map $X$. The query mapping function $f$ maps the input feature map $X$ to a lower-dimensional query $q \in \mathbb{R}^{K}$ as follows:
% \begin{equation}
%     \label{eqn:query_mapping}
%     q = f(X),
% \end{equation}
% This step  transforms the feature map $X$ of any size into a query $q$ of the dimensionality $K$. 
% % transforming feature map $X$ of any size into a query $q$ with dimensionality $K$. 

% \subsubsection{2) Expert Retrieval}
\noindent\textbf{2) Expert Retrieval} 
% In this part, we show how we can retrieve the most suitable expert from each Expert Bank, taking the $p$-th Expert Bank as an example.
We now show how we can retrieve the most suitable expert from each Expert Bank for input $X$, taking the $p$-th Expert Bank as an example. %%based on input X
As shown in Fig.~\ref{fig:deer}, we first extract a compact and meaningful representation from the input feature map $X$. The query mapping function $f^{p}$ maps the input feature map $X$ to a lower-dimensional query $q^{p} \in \mathbb{R}^{K}$ as follows:
\begin{equation}
    % \vspace{-1cm}
    \label{eqn:query_mapping}
    q^{p} = f^{p}(X).
\end{equation}

% This step transforms the feature map $X$ of any size into a query $q^{p}$ of the dimensionality $K$. 
This step transforms the feature map $X$ into a query (vector) $q^{p}$ of the dimensionality $K$.

% Next, conditioned on $q$, we would like to retrieve the $d$ most suitable experts from the Expert Banks for discriminating subtle cues within the feature map $X$. Recall that each expert $E_{i}^{p}$ holds a key $k_{i}^{p}$ which represents its area of expertise, while $q$ is the representation of the feature map $X$. The degree of suitability of the expert $E_{i}^{p}$ on the feature map $X$ can thus be obtained by calculating the matching score between $k_{i}^{p}$ and $q$. 
% For the $p$-th Expert Bank, we calculate the matching score $s_{i}^{p}$ for each expert $E_{i}^{p}$ using dot product between the query $q$ and the key $k_{i}^{p}$ as:

% Next, conditioned on $q$, we would like to retrieve the most suitable expert from each Expert Bank for discriminating subtle cues within the feature map $X$. 
% Next, conditioned on $q$, we now show how we retrieve the most suitable expert from an Expert Bank for discriminating subtle cues within the feature map $X$. 
% Next, we show how we retrieve the most suitable expert, conditioned on $q$, from an Expert Bank for discriminating subtle cues within the feature map $X$. 
% Next, taking the $p$-th Expert Bank as an example, we show how we can retrieve the most suitable expert, conditioned on $q$, for discriminating subtle cues within the feature map $X$. 
% Next, conditioned on $q^{p}$, we would like to retrieve the most suitable expert from the $p$-th Expert Bank for discriminating subtle cues within the feature map $X$. 
Next, conditioned on $q^{p}$, we retrieve the most suitable expert from the $p$-th Expert Bank for discriminating subtle cues within the feature map $X$. 
Recall that each expert $E_{i}^{p}$ holds a key $k_{i}^{p}$ which represents its area of expertise, while $q^{p}$ is the representation of the feature map $X$. The degree of suitability of the expert $E_{i}^{p}$ on the feature map $X$ can thus be obtained by calculating the matching score between $k_{i}^{p}$ and $q^{p}$. 
% For each Expert Bank, we calculate the matching score $s_{i}$ for each expert $E_{i}$ using dot product between the query $q$ and the key $k_{i}$ as:
We calculate the matching score $s_{i}^{p}$ for each expert $E_{i}^{p}$ using dot product between the query $q^{p}$ and the key $k_{i}^{p}$ as:

% \vspace{-1em}
\begin{align}
    \label{eqn:score_calc}
    &s_{i}^{p} = q^{p \top} k_{i}^{p}, ~~~ i = \{1,2,...,M \}. \\
    % &s_{i}^{p} = q^\top k_{i}^{p}, i = \{1,2,...,M \}. \\
    \label{eqn:argmax}
    &I^{p} = Argmax_i (\{ s_{i}^{p} \}_{i=1}^M),
% \vspace{-0.15cm}
\end{align}
%
% \lgg{need to polish this part more....}
% where $Argmax_i$ returns the index $i$ belonging to the largest element in the set $\{ s_{i}^{p} \}_{i=1}^M$.
% $I^p$ in Eq.~\ref{eqn:argmax} will then represent the index of the retrieved expert.
where $Argmax_i$ returns the index $i$ belonging to the largest element in the set $\{ s_{i}^{p} \}_{i=1}^M$, and the returned index $I^p$ represents the index of the retrieved expert.
% $I^p$ in Eq.~\ref{eqn:argmax} will then represent the index of the retrieved expert.
%where $Argmax_i$ returns the index $i$ belonging to the largest element in the set.
% As Eq.~\ref{eqn:score_calc}
% The higher the matching score, the closer the feature map is, to the area of expertise represented by the key.
% This is because, the highest matching score within the set ($s_{I^p}^{p}$) will come from the key $k_{I^p}^{p}$ that matches the query $q$ the most.
% This is because(\ke{i dont understand why there is this is because here, maybe remove?}), 
We take the highest matching score ($s_{I^p}^{p}$) within the set, as it will come from the key $k_{I^p}^{p}$ representing an area of expertise that matches the query $q^{p}$ the best.
% The highest matching score within the set ($s_{I^p}^{p}$) will come from the key $k_{I^p}^{p}$ representing an area of expertise that matches the query $q$ the most.
Thus, the corresponding expert $E_{I^p}^{p}$ is the most suitable expert to be applied to the feature map $X$, and is retrieved from the $p$-th Expert Bank in this step.

It is worth mentioning that, using this key-query mechanism, the input feature maps that are highly similar (i.e., with similar $q$ values) %will have similar queries and tend to
will tend to have high matching scores with the same key and retrieve the same expert. Crucially, this leads to the experts having to discriminate between highly similar input samples, pushing each expert to specialize in exploiting subtle cues for distinguishing between those similar samples to tackle early action prediction.
% Notably, we conduct this Expert Retrieval operation throughout the $d$ Expert Banks in a \textit{parallel} and \textit{efficient} manner.
% We note that, although only the operation for the $p$-th Expert Bank is shown in Eq.~\ref{eqn:score_calc},\ref{eqn:argmax}, it can be conducted for all $d$ Expert Banks in a \textit{parallel} and \textit{efficient} manner.
% Notably, we conduct this Expert Retrieval operation throughout the $d$ Expert Banks in a \textit{parallel} and \textit{efficient} manner.
% As a remark on efficiency,
% In terms of efficiency, we remark that operations shown in Eq.~\ref{eqn:score_calc},\ref{eqn:argmax} can be conducted for all $d$ Expert Banks in a \textit{parallel} and \textit{efficient} manner.

% Above, we retrieve experts from the $p$-th Expert Bank, but the same process is conducted for all $d$ Expert Banks to retrieve $d$ experts.
Above, we only show the operations on the $p$-th Expert Bank, but the same process is conducted for all $d$ banks to retrieve $d$ experts, which is shown in Fig.~\ref{fig:deer}.
Notably, this process (Eq.~\ref{eqn:query_mapping}, \ref{eqn:score_calc}, \ref{eqn:argmax}) across $d$ Expert Banks can be done in \textit{parallel}, so it is \textit{efficient}.
% In terms of efficiency, we remark that operations shown in Eq.~\ref{eqn:query_mapping},\ref{eqn:score_calc},\ref{eqn:argmax} can be conducted for all $d$ Expert Banks in a \textit{parallel} and \textit{efficient} manner.

% \lgg{define the new notation here.}
% \noindent\textbf{Expert Assembly.} To construct the expert block $W_{expert}$, we assemble the retrieved expert kernels from all $d$ Expert Banks (i.e., $\{m_{I^p}^p\}_{p=1}^{d}$), as follows: 
% \noindent\textbf{Expert Assembly.} We assemble the retrieved expert kernels from all $d$ Expert Banks (i.e., $\{m_{I^p}^p\}_{p=1}^{d}$) to form an expert block $W_{expert}$ as follows:
% To construct the expert block $W_{expert}$, we assemble the retrieved expert kernels from all $d$ Expert Banks (i.e., $\{m_{I^p}^p\}_{p=1}^{d}$), as follows: 
% \subsubsection{3) Expert Assembly}
\noindent\textbf{3) Expert Assembly} 
% In the Expert Retrieval step, we retrieve $d$ expert kernels (i.e., $\{m_{I^p}^p\}_{p=1}^{d}$) from $d$ Expert Banks.
We assemble the retrieved expert kernels from all $d$ Expert Banks (i.e., $\{m_{I^p}^p\}_{p=1}^{d}$) to form an expert block $W_{expert}$ as follows:
\begin{align}
    \label{concat}
    % W_{expert} = Concat(\{m_i\}_{i \in \mathcal{I}}),
    W_{expert} = Concat(\{m_{I^p}^p\}_{p=1}^{d}),
\end{align}
% where $W_{expert}$ 
where $W_{expert} \in \mathbb{R}^{d \times N_{in} \times b_h \times b_w}$ is composed of the parameters of the $d$ experts that have the highest matching scores in the $d$ banks, and $Concat$ denotes concatenation along the channel dimension. % (in increasing order of $n$)
% These retrieved experts will be specialized in feature maps that are similar to $X$ (but may belong to other classes). This encourages the experts to capture %and will be more sensitive to 
% subtle cues that distinguish between the true class of $X$ and other similar classes for tackling early action prediction. 
% These retrieved experts will be specialized in feature maps that are similar to $X$ (but may belong to other classes). This encourages the experts to capture
% multiple subtle cues that distinguish between the true class of $X$ and other similar classes for tackling early action prediction. 
These retrieved experts will be specialized in capturing multiple subtle cues in $X$, that distinguish between the true class of $X$ and other similar classes for tackling early action prediction.

% Finally, to form the full convolutional block $W_{all}$, we further concatenate the non-expert block $W_{nonexpert}$ and the expert block $W_{expert}$ as:
Finally, to form the full convolutional block $W_{ERA}$, we further assemble the non-expert block $W_{nonexpert}$ and the expert block $W_{expert}$ (shown in Fig.~\ref{fig:deer}) as:
\begin{equation}
    W_{ERA} = Concat(W_{nonexpert},W_{expert}).
\end{equation}

The final assembled block $W_{ERA}$ will be applied to the feature map in a similar manner to a traditional convolutional kernel $W_{conv}$. 
Notably, as $W_{ERA}$ can directly replace $W_{conv}$, our ERA module is a plug-and-play module that can replace the basic convolutional layer.
% Notably, our ERA module is a plug-and-play module that can replace the basic convolutional layer.
% , and also possesses improved discriminative power by utilizing both non-expert and expert parameters.

To apply the ERA module to other types of convolutions that are used in early action prediction architectures, only minor changes need to be made. For 1D and 3D convolutions, we change the shape of $m_{i}^{p}$ and $W_{nonexpert}$ according to the corresponding 1D or 3D kernel. As for graph convolutions, since many graph convolutions (as used in \cite{shi2019two}) are implemented based on traditional convolutions with additional parameters and steps to account for adjacency information, thus we can also implement our method by replacing the contained convolutional kernel with our ERA module in these scenarios.

\noindent{\textbf{4) Analysis of specialization of experts}} 
Next, we analyze how our ERA module allows experts to specialize in subtle cues through the expert-retrieval mechanism during training. 
% This justification is rather important, as it explains why our retrieval of the most suitable expert in each Expert Bank tackles the sub-optimal training behaviour of the deep neural networks, leading them to go beyond learning general patterns towards better discrimination of subtle differences. 
This justification is rather important, as it explains why the retrieval of the most suitable $d$ experts leads to better discrimination of subtle differences and tackles the sub-optimal training behaviour of the deep neural networks. 
We approach this by analyzing the
differences between the gradients that update experts and non-experts \textit{during backpropagation}. 
% when the ERA module (as described in the previous part) is used.

Usually, the aggregated gradients $\bar{g}$ of a loss $\mathcal{L}$ w.r.t a model parameter $w$ is computed by averaging over the entire batch with batch size $B$:
\begin{equation}
\label{eqn:normal_grad}
    \bar{g} = \frac{1}{B} \sum_{j=1}^B \frac{\partial \mathcal{L}_j}{\partial w}.
% \vspace{-0.2cm}
\end{equation}

% In a static network, all parameters are updated using $\bar{g}$ in Eq.~\ref{eqn:normal_grad}, which trains the parameters to contribute towards classifying all samples, resulting in the learning of general patterns that apply to more samples, as opposed to subtle differences that occur only in a small subset of the data.
The non-expert parameters are updated using $\bar{g}$ in Eq.~\ref{eqn:normal_grad}, which trains the non-expert parameters to contribute towards classifying all samples, resulting in the learning of general patterns that apply to more samples, as opposed to subtle differences that occur only in a small subset of the data.
If all parameters in a network are non-experts, this results in the network having sub-optimal performance with respect to subtle cues \cite{johnson2019survey} and leads to worse performance on early action prediction.
% If all parameters in a network are non-experts, this results in ``lazy" behaviour where the network has sub-optimal performance with respect to subtle cues \cite{johnson2019survey} and leads to worse performance on early action prediction.
% If all parameters in a network are non-experts, this results in ``lazy" behaviour where the network has sub-optimal performance with respect to subtle cues \cite{johnson2019survey} and leads to worse performance on early action prediction.
% Such sub-optimal training behaviour with respect to subtle cues \cite{johnson2019survey} leads to worse performance on early action prediction.

% In contrast, in our ERA module, not all experts are selected by each sample, as each expert is only retrieved for its most suitable samples. 
% When backpropagating using the loss $\mathcal{L}$ on experts, the aggregated gradient $\bar{g}_{i}^{p}$ for the expert kernel weights $m_{i}^{p}$ thus becomes: 
% \begin{equation}
% \label{eqn:expert_grad}
%     \bar{g}_{i}^{p} = \frac{1}{|\mathcal{N}(k_{i}^{p})|} \sum_{j \in \mathcal{N}(k_{i}^{p})} \frac{\partial \mathcal{L}_j}{\partial m_{i}^{p}},
% % \vspace{-0.2cm}
% \end{equation}
% where $\mathcal{N}(k_{i}^{p})$ denotes the set of samples in the mini-batch that select expert $E_{i}^{p}$ (with key $k_{i}^{p}$ and kernel $m_{i}^{p}$) as one of their $d$ experts, i.e., $\mathcal{N}(k_{i}^{p}) = \{ j \text{ s.t } I_j^p = i \}_{j=1}^{B}$, 
% where $I_j^p$ refers to the index of the selected expert in the $p$-th Expert Bank ($I^p$) for the $j$-th sample.

In contrast, in our ERA module, not all experts are selected by each sample, as each expert is only retrieved for its most suitable samples. 
When backpropagating using the loss $\mathcal{L}$ on experts, the aggregated gradient $\bar{g}_{i}^{p}$ for the expert kernel weights $m_{i}^{p}$ thus becomes: 
\begin{equation}
\label{eqn:expert_grad}
    \bar{g}_{i}^{p} = \frac{1}{|\mathcal{N}(k_{i}^{p})|} \sum_{j \in \mathcal{N}(k_{i}^{p})} \frac{\partial \mathcal{L}_j}{\partial m_{i}^{p}},
% \vspace{-0.2cm}
\end{equation}
% where $\mathcal{N}(k_{i})$ denotes the set of samples in the mini-batch that select expert $E_{i}$ (with key $k_{i}$ and kernel $m_{i}$) as one of their $d$ experts, i.e., $\mathcal{N}(k_{i}) = \{ j \text{ s.t } I_j = i \}_{j=1}^{B}$,
where $\mathcal{N}(k_{i}^{p})$ denotes the set of samples in the batch that select expert $E_{i}^{p}$ (with key $k_{i}^{p}$ and kernel $m_{i}^{p}$), i.e., $\mathcal{N}(k_{i}^{p}) = \{ j \text{ s.t } I_j^{p} = i \}_{j=1}^{B}$,
where $I_j^{p}$ refers to the index of the selected expert in the $p$-th Expert Bank ($I^{p}$) for the $j$-th sample in the batch.
The samples in $\mathcal{N}(k_{i}^{p})$ are likely to be very similar, with only some subtle differences, due to their close proximity to $k_{i}^{p}$ in the feature space.

%\textcolor{red}{again referring to Eq.7. Why in such a way???? it looks you are jumping from one to the other in several places and it makes it confusing. } 
If we train the expert $E_{i}^{p}$ using the gradient $\bar{g}_{i}^{p}$ as in Eq.~\ref{eqn:expert_grad}, the expert is only updated using samples that are closer to this expert's area of expertise, i.e., samples which are in $\mathcal{N}(k_{i}^{p})$. 
% Due to their close proximity to $k_{i}$ in the feature space, these samples are likely to be very similar with only some subtle differences. 
% By focusing only on a subset of samples that are highly similar, $\bar{g}_{i}$ pushes the expert to learn to discriminate them through their subtle differences. 
% As $\bar{g}_{i}$ does not include samples outside of $\mathcal{N}(k_{i})$, it does not encourage the expert $E_{i}$ to learn general patterns that generally hold across all data, and instead, \textit{pushes the expert to learn to exploit subtle differences} among the samples in $\mathcal{N}(k_{i})$ to make them more distinguishable.
% Thus, by training using $\bar{g}_{i}$ as compared to $\bar{g}$, more emphasis is placed on learning to distinguish between similar samples, as expert $E_i$ is pushed to learn to exploit subtle differences among the samples in $\mathcal{N}(k_{i})$.
% Thus, as compared to $\bar{g}$, much more emphasis is placed on learning to distinguish between similar samples (in $\mathcal{N}(k_{i})$ only), which \textit{pushes the expert to learn to exploit subtle differences} as opposed to general patterns that generally hold across all data.
Thus, as compared to $\bar{g}$, much more emphasis is placed on learning to distinguish between these similar samples in $\mathcal{N}(k_{i}^{p})$ only, which \textit{pushes the expert to learn to exploit subtle differences in these samples}, as opposed to general patterns that generally hold across all data. 
%%%%%% deleted from camera ready %%%%%%%%%%%%%
%%This is also validated qualitatively in our filter visualizations (placed in the Supplementary), which show that experts encode sharper patterns compared to non-experts.

% \subsection{}

\subsection{Expert Learning Rate Optimization} 
% \lgg{to rewrite and change notation?}
% Experts in our ERA modules, together with all other network parameters, are end-to-end trainable using backpropagation.
% However, due to the uneven distribution of samples across experts, some experts might be selected by more samples and be better trained than others, possibly causing imbalanced training that limits performance of our ERA module. 
% To mitigate this effect, we design an Expert Learning Rate Optimization method that optimizes the training among experts, leading to improved early action prediction accuracy.
% We introduce expert learning rates $\beta$ as additional parameters, where each element 
% $\beta_{i}^{p}$ corresponds to an expert $E_{i}^{p}$ and is initialized to $1$. 
% Instead of using manual tuning to adjust the large set (all experts in all ERA modules) of $\beta$, we update $\beta$ using a \textit{meta-learning approach} in an end-to-end manner.
Experts in our ERA modules, together with all other network parameters, are end-to-end trainable using backpropagation.
However, due to the uneven distribution of samples across experts, some experts might be selected by more samples and be better trained than others, possibly causing imbalanced training that limits the performance of our ERA module. 
To mitigate this effect, we design an Expert Learning Rate Optimization (ELRO) method that optimizes the training among experts, leading to improved early action prediction accuracy.
% We introduce expert learning rates $\beta$ as additional parameters, where each element 
% $\beta_{i}^{p}$ corresponds to an expert $E_{i}^{p}$ and is initialized to $1$. 
% We introduce a set of expert learning rates $\beta = \{\beta_{i}^{p}\}_{i=1,p=1}^{i=M,p=d}$ as additional parameters, where each element 
% We introduce a set of expert learning rates $\beta = \{\beta_{i}^{p}\}_{i=1,p=1}^{M,d}$ as additional parameters, where each element 
% \lgg{for clarity, we only show one ERA module in this section...}
% For ease of notation, in this section, we assume that we only have one ERA module, although this method can also work for multiple ERA modules.
For ease of notation, in this section, we only use one ERA module, although this method can also work for multiple ERA modules.
We introduce a set of expert learning rates $\beta = \{\beta_{i}^{p}\}_{i \in \{1..M \},p \in \{1..d \}}$ as additional parameters, where each element 
$\beta_{i}^{p}$ is a scalar that balances the training of a corresponding expert $E_{i}^{p}$ during backpropagation. % and is initialized to $1$ 
% Instead of using manual tuning to adjust the large set (all experts in all ERA modules) of $\beta$, we update $\beta$ using a \textit{meta-learning approach} in an end-to-end manner.
Instead of using manual tuning to adjust the large set of $\beta$, we update $\beta$ using a \textit{meta-learning approach} in an end-to-end manner.

% The core idea of meta-learning \cite{finn2017maml,shu2019metaweightnet} is about learning-to-learn, which in our case is learning (through feedback from a validation set) to\textit{ optimize the learning rate} $\beta_{i}^{p}$ for each expert $E_{i}^{p}$ for \textit{improved training of experts}.
% The core idea of meta-learning \cite{finn2017maml,shu2019metaweightnet} is about learning-to-learn, which in our case is \textit{learning to optimize the learning rate} $\beta_{i}^{p}$ for each expert $E_{i}^{p}$ for \textit{improved training of experts}.
The core idea of meta-learning \cite{finn2017maml,shu2019metaweightnet} is about ``learning-to-learn'', which in our case is \textit{learning to optimize the learning rates} $\beta$ for \textit{improved training of experts}.
% To achieve this, we first simulate training on a training set using the current $\beta$ values, and then evaluate the improvement of that simulated training step on a validation set.
% This meta-optimization of each $\beta_{i}^{p}$ is conducted over two steps.
This meta-optimization of $\beta$ is conducted over two steps.
% Firstly, we simulate training on a training set while using the current $\beta$ values to balance expert updates, to obtain a virtually updated \textit{interim model}.
% Firstly, we simulate training on a training set while using the current $\beta_{i}^{p}$ values to balance updates for each expert $E_{i}^{p}$, to obtain a virtually updated \textit{interim model}.
Firstly, we simulate training on a training set while using the current $\beta_{i}^{p}$ values to balance updates for each expert $E_{i}^{p}$ respectively, to obtain a virtually updated \textit{interim model}.
% Firstly, we simulate training on a training set while balancing updates for each expert $E_{i}^{p}$ using $\beta_{i}^{p}$, to obtain a virtually updated \textit{interim model}.
% Next, we evaluate the performance of the interim model on a validation set, such that the gradients of these losses provide feedback on how we can adjust $\beta$ to improve the effectiveness of the simulated training step on the unseen validation samples (which signals generalizable improvement).
% Next, we evaluate the performance of the interim model on a validation set, and the gradients of these losses provide feedback on how we can adjust $\beta$ to a more optimized $\beta'$, to improve the effectiveness of the simulated training step on the unseen validation samples (which signals generalizable improvement).
% Next, we evaluate the performance of this interim model on a validation set, and the gradients of these validation losses will \textit{provide feedback on how we can adjust $\beta$ to a more optimized $\beta'$}, such that training results in better performance on unseen validation samples (which signals generalizable improvement).
Next, we evaluate the performance of this interim model on a validation set, and the gradients of these validation losses will \textit{provide feedback on how we can adjust $\beta$ to a more optimized $\beta'$} (which improves training of experts, and results in better performance on unseen validation samples).
% (using which improves training of experts and results in better performance on unseen validation samples).
% The second-order gradients from the losses on the validation set provide feedback about how we can adjust $\beta$ to improve the effectiveness of the simulated training step on the unseen validation samples (which signals generalizable improvement). 
% The second-order gradients from the losses on the validation set provide feedback about how we can adjust $\beta$ to improve the effectiveness of the simulated training step on the unseen validation samples (which signals generalizable improvement). 
% After meta-optimizing the $\beta$ values using the second-order gradients, our actual training step that updates experts will lead to greater improvements in performance.
% Using the meta-optimized $\beta'$ values for training experts will then lead to greater improvements in performance.
% Finally, we use the meta-optimized $\beta'$ values for training experts, which will then lead to greater improvements in performance.
Finally, we then use the meta-optimized $\beta'$ values for balancing expert updates during actual model training, which yields improvements in performance.

% \lgg{highlight that fixed beta is used}
% \lgg{highlight the temporary middle model}
% \lgg{highlight that the temporary middle model provides feedback to adjust beta}
% \lgg{maybe can try to think of a better name than interim model?}

% The core idea of meta-learning \cite{finn2017maml,shu2019metaweightnet} is about learning-to-learn, which in our case is learning (through second-order updates on a validation set) to\textit{ optimize the learning rate} $\beta_{i}^{p}$ for each expert $E_{i}^{p}$ for \textit{improved training of experts}.
% %% through updating on a validation set, to learn to optimize $\beta$ for better training
% To achieve this, we first simulate training on a training set using the current $\beta$ values, and then evaluate the improvement of that simulated training step on a validation set.
% The second-order gradients from the losses on the validation set provide feedback about how we can adjust $\beta$ to improve the effectiveness of the simulated training step on the unseen validation samples (which signals generalizable improvement). 
% % After meta-optimizing of the $\beta$ values using the second-order gradients, our actual training step that updates experts will lead to greater improvements in performance.
% After meta-optimizing the $\beta$ values using the second-order gradients, our actual training step that updates experts will lead to greater improvements in performance.

\begin{figure*}
    \centering
    \includegraphics[width=\linewidth]{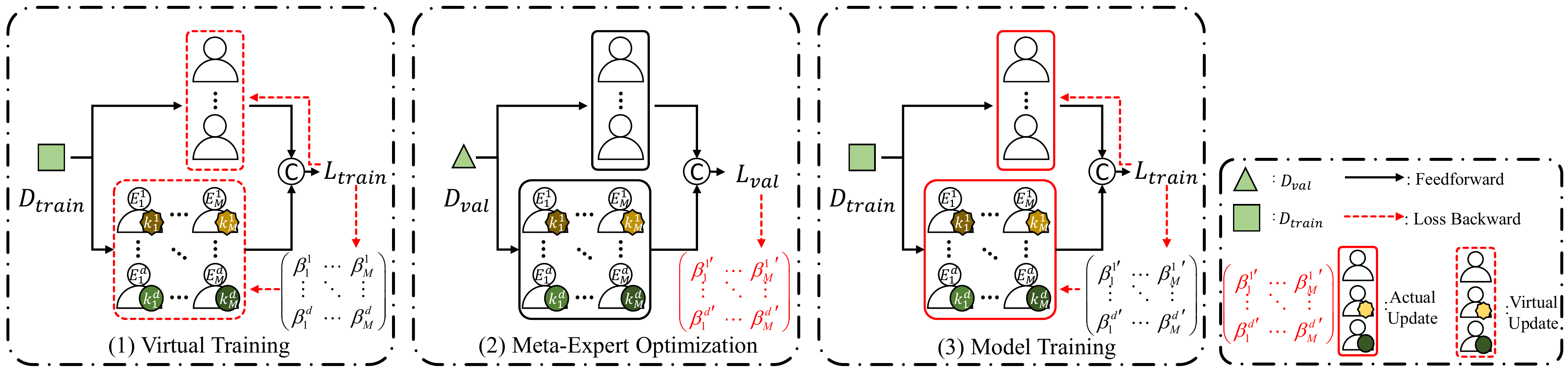}
    % \vspace{-0.75cm}
    \caption{
    Illustration of the Expert Learning Rate Optimization method. 
    % Illustration of the ELRO method.     
    Two independent batches are sampled: training samples $\mathcal{D}_{train}$ and validation samples $\mathcal{D}_{val}$. Forward propagation paths are in black, while backpropagation paths are in red. The entire method consists of three phases: (1) Virtual Training, (2) Meta-Expert Optimization and (3) Model Training. At the Virtual Training step, all non-$\beta$ model parameters are virtually updated using $\mathcal{D}_{train}$. 
    % At the Meta-Expert Optimization step, expert learning rate parameters $\beta$ are dynamically updated using the second-order gradients from validation loss $\mathcal{L}_{val}$. 
    At the Meta-Expert Optimization step, expert learning rate parameters $\beta$ are dynamically updated using the gradients from validation loss $\mathcal{L}_{val}$.     
    At the Model Training step, all non-$\beta$ parameters are updated using $\mathcal{D}_{train}$ with updated $\beta'$.
    Best viewed in color. 
    % \lgg{update to show multiple expert banks in the method}
    }
    \label{fig:moe}
% \vspace{-0.5cm}
\end{figure*}

An illustration of our proposed ELRO method is shown in Fig \ref{fig:moe}. Specifically,
in each iteration, we draw two batches of training data 
% \textcolor{green}{is it better to say "two mini-batches of training data"? just to avoid any confusion} 
that are non-overlapping, which we call training samples $\mathcal{D}_{train}$ and validation samples $\mathcal{D}_{val}$.
Then, the following three steps are employed to update the model parameters.

% \textit{(1) Virtual Training.}
% We simulate the training on $\mathcal{D}_{train}$ by virtually updating all the model parameters other than $\beta$ as follows:
% \begin{align}
%   \hat{w} = w- \alpha  \nabla_{w} \mathcal{L} (w,E;\mathcal{D}_{train}), \\
%   \label{eqn:virtual_train_expert}
%   \hat{E} = E- \alpha \beta \nabla_E \mathcal{L} (w,E;\mathcal{D}_{train}),
% \end{align}
% where $w$ represents the non-expert parameters, $\alpha$ is the learning rate (which is 
% a fixed hyperparameter), and $\mathcal{L}$ refers to the supervised loss for the early action prediction task. Note that here the updates of experts $E$ (which includes keys $k$ and kernel weights $m$) are further scaled by $\beta$. 

\textit{(1) Virtual Training.}
We simulate the training on $\mathcal{D}_{train}$ by virtually updating all the model parameters other than $\beta$ as follows:
\begin{align}
  \hat{w} &= w- \alpha  \nabla_{w} \mathcal{L} (w,\mathcal{E};\mathcal{D}_{train}), \\
  \label{eqn:virtual_train_expert}
  \hat{E}_{i}^{p} = E_{i}^{p} - \beta_{i}^{p} \nabla_{E_{i}^{p}} &\mathcal{L} (w,\mathcal{E};\mathcal{D}_{train}),
  i = \{1,..,M \}, p = \{1,..,d \},
\end{align}
% where $w$ represents the non-expert parameters, $\alpha$ is the learning rate (which is 
% a fixed hyperparameter), and $\mathcal{L}$ refers to the supervised loss for the early action prediction task. 
% Note that here the updates of experts $E$ (which includes keys $k$ and kernel weights $m$) are further scaled by $\beta$.
where $w$ represents the non-expert parameters, $\mathcal{E}$ represents the set of experts $\{E_{i}^{p}\}_{i \in \{1..M \},p \in \{1..d \}}$, $\alpha$ is the learning rate (which is 
a fixed hyperparameter), and $\mathcal{L}$ refers to the supervised loss for the early action prediction task. 
Note that here the update of each expert $E_{i}^{p}$ (which includes key $k_{i}^{p}$ and kernel weights $m_{i}^{p}$) is scaled by $\beta_{i}^{p}$.

% \textit{(2) Meta-Expert Optimization.}
% % In this step, we evaluate the performance of the virtually updated model, $\hat{w}$ and $\hat{E}$, on $\mathcal{D}_{val}$.
% In this step, we evaluate the performance of the virtually updated model, $\hat{w}$ and $\hat{\mathcal{E}}$, on $\mathcal{D}_{val}$.
% The gradients with respect to $\beta$ provide feedback on how $\beta$ should be tuned to improve the learning process to generalize better to unseen samples, as follows:
% \begin{equation}
%   \label{eqn:metagradient}
%   \beta' = \beta - \alpha  \nabla_{\beta} \mathcal{L} (\hat{w},\hat{E};\mathcal{D}_{val}). 
% % \vspace{-0.2cm}
% \end{equation}
% Only $\beta$ is updated in this step, and other parameters ($\hat{w}$ and $\hat{E}$) remain fixed. 
% Note that $\nabla_{\beta}$ takes gradients with respect to $\beta$ as used in Eq.~\ref{eqn:virtual_train_expert}.
% By tuning $\beta$ with the meta-gradients in Eq.~\ref{eqn:metagradient}, the newly updated learning rates $\beta'$ of experts can provide better training for the experts.   

\textit{(2) Meta-Expert Optimization.}
In this step, we evaluate the performance of the virtually updated model (consisting of $\hat{w}$ and $\hat{\mathcal{E}}$) on $\mathcal{D}_{val}$.
The gradients w.r.t each expert learning rate $\beta_{i}^{p}$ provide feedback on how $\beta_{i}^{p}$ should be tuned for the virtually updated model to generalize better to unseen samples, as follows:
\begin{equation}
  \label{eqn:metagradient}
  \beta_{i}^{p}{}' = \beta_{i}^{p} - \alpha  \nabla_{\beta_{i}^{p}} \mathcal{L} (\hat{w},\hat{\mathcal{E}};\mathcal{D}_{val}),  
  i = \{1,..,M \}, p = \{1,..,d \}.
% \vspace{-0.2cm}
\end{equation}
Only $\beta$ is updated in this step, and other parameters ($\hat{w}$ and $\hat{\mathcal{E}}$) remain fixed. 
Note that $\nabla_{\beta_{i}^{p}}$ takes gradients with respect to $\beta_{i}^{p}$ as used in Eq.~\ref{eqn:virtual_train_expert}.
% By tuning $\beta_{i}^{p}$ with the meta-gradients in Eq.~\ref{eqn:metagradient}, the newly updated expert learning rate $\beta_{i}^{p}'$ can provide better training for expert $E_{i}^{p}$.  
% This means that, by tuning $\beta_{i}^{p}$ in Eq.~\ref{eqn:metagradient}, the newly updated expert learning rate $\beta_{i}^{p}'$ can provide better training for expert $E_{i}^{p}$.  
This means that, by tuning $\beta_{i}^{p}$ in Eq.~\ref{eqn:metagradient}, the newly updated expert learning rate $\beta_{i}^{p}{}'$ can provide better training for expert $E_{i}^{p}$ if Eq.~\ref{eqn:virtual_train_expert} is performed again.

% \textit{(3) Model Training.}
% % Model parameters, $E$ and $w$, are updated using the meta-optimized $\beta'$ on $\mathcal{D}_{train}$ as:
% After we get meta-optimized $\beta'$, we can update model parameters $E$ and $w$ on $\mathcal{D}_{train}$ as:
% \begin{align}
%   w' = w- \alpha  \nabla_{w} \mathcal{L} (w,E;\mathcal{D}_{train}), \\
%   E' = E- \alpha \beta' \nabla_E \mathcal{L} (w,E;\mathcal{D}_{train}).
% \end{align}
% % In this step, the dynamically adjusted $\beta'$ balances the training of experts to benefit more from training, and leads to improved performance.
% In this step, the meta-optimized $\beta'$ \textit{balances the training of experts such that performance on unseen samples is improved}.
% % Above depicts one iteration of Expert Learning Rate Optimization, where we have obtained updated parameters $w'$, $E'$ and $\beta'$.
% This concludes one iteration of Expert Learning Rate Optimization, where we have obtained updated parameters $w'$, $E'$ and $\beta'$.
% An outline of this algorithm is shown in the Supplementary Material.

\textit{(3) Model Training.}
After we obtain the set of meta-optimized expert learning rates $\beta' = \{\beta_{i}^{p}{}'\}_{i \in \{1..M \},p \in \{1..d \}}$, we can perform actual model training by updating model parameters $\mathcal{E}$ and $w$ on $\mathcal{D}_{train}$ as:
% \vspace{-0.2cm}
\begin{align}
   w' &= w- \alpha  \nabla_{w} \mathcal{L} (w,\mathcal{E};\mathcal{D}_{train}), \\
  E_{i}^{p}{}' = E_{i}^{p} -  \beta_{i}^{p}{}' \nabla_{E_{i}^{p}} &\mathcal{L} (w,\mathcal{E};\mathcal{D}_{train}),
  i = \{1,..,M \}, p = \{1,..,d \},
\end{align}
% In this step, the dynamically adjusted $\beta'$ balances the training of experts to benefit more from training, and leads to improved performance.
In this step, the meta-optimized $\beta'$ \textit{balances the training of experts such that performance on unseen samples is improved}.
% Above depicts one iteration of Expert Learning Rate Optimization, where we have obtained updated parameters $w'$, $E'$ and $\beta'$.
This concludes one iteration of ELRO, where we have obtained updated parameters $w'$, $\mathcal{E}'$ and $\beta'$.
An outline of this algorithm is shown in the Supplementary Material.

\subsection{Loss function}
% \vspace{-0.1cm}
\label{sec:loss}
% During training, our loss $\mathcal{L}$ is given by $\mathcal{L}=  \mathcal{L}_{CE} - \gamma_{s} \mathcal{L}_{s}$, 
% where $\mathcal{L}_{CE}$ is the cross-entropy loss on the early action prediction task and $\gamma_{s}$ is a hyperparameter that weights the relative significance of losses. 
% $\mathcal{L}_{s}$ refers to a similarity loss that we implement between the expert kernels in each Expert Bank.
% It encourages more diverse experts by penalizing experts that get too close to each other, and we find that it brings some improvement in practice. 
% % We also observe the impact of this loss qualitatively, and visualizations have been placed in the Supplementary Material. 
% We implement it using a negative pairwise mean-squared loss among expert kernels, 
% i.e., $\mathcal{L}_{s} = \sum_{p=1}^d \sum_{i=1}^M \sum_{j \neq i} ||m_i^{p} - m_j^{p}||^2$ for each individual ERA module.
% The loss is applied on all ERA modules in our network.
% \lgg{focus on the fact that we actually only need CE. just add similarity loss for good measure...}

We train our model using a cross-entropy loss $\mathcal{L}_{CE}$ on the early action prediction task.
Furthermore, we find that applying an additional similarity loss $\mathcal{L}_{s}$ brings some improvements in practice, where
the similarity loss $\mathcal{L}_{s}$ penalizes experts that get too close to each other, which encourages experts to be more diverse.
Specifically, we implement $\mathcal{L}_{s}$ on all ERA modules within our network, using a negative pairwise mean-squared loss among expert kernels in each Expert Bank,
% i.e., $\mathcal{L}_{s} = \sum_{p=1}^d \sum_{i=1}^M \sum_{j \neq i} ||m_i^{p} - m_j^{p}||^2$ for each individual ERA module.
i.e., using one ERA module as an example, $\mathcal{L}_{s} = \sum_{p=1}^d \sum_{i=1}^M \sum_{j \neq i} ||m_i^{p} - m_j^{p}||^2$.
% Using $\mathcal{L}_{s}$.
Overall, our loss $\mathcal{L}$ is then given by $\mathcal{L}=  \mathcal{L}_{CE} - \gamma_{s} \mathcal{L}_{s}$,
where $\gamma_{s}$ is a hyperparameter that weights the relative significance of losses. 
% $\mathcal{L}_{s}$ refers to a similarity loss that we implement between the expert kernels in each Expert Bank.
% It encourages more diverse experts by penalizing experts that get too close to each other, and we find that it brings some improvement in practice. 
% We also observe the impact of this loss qualitatively, and visualizations have been placed in the Supplementary Material. 
% We implement $\mathcal{L}_{s}$ using a negative pairwise mean-squared loss among expert kernels, 
% i.e., $\mathcal{L}_{s} = \sum_{p=1}^d \sum_{i=1}^M \sum_{j \neq i} ||m_i^{p} - m_j^{p}||^2$ for each individual ERA module.
% The loss is applied on all ERA modules in our network.
% \lgg{focus on the fact that we actually only need CE. just add similarity loss for good measure...}

% \begin{align}
%     % \mathcal{L}_{similarity} = \sum_{r=1}^R \sum_{p=1}^d \mathcal{L}_{similarity,r,p}  \\
%     \mathcal{L}_{s} = \sum_{p=1}^d \sum_{i=1}^M \sum_{j \neq i} \text{Mean-Square} (m_i^{p}, m_j^{p})
% \end{align}
% % where there are $R$ ERA modules throughout the network,
% % and $m_i^{r,p}$ refers to the $i$-th expert kernel in the $p$-th partition of the $r$-th ERA module. 

% \vspace{-0.1cm}
\section{Experiments}
% \vspace{-0.1cm}
% \subsection{Datasets}

% We conduct extensive experiments on both skeletal and RGB datasets to validate the effectiveness of our proposed ERA module for early action prediction.
To validate effectiveness of our ERA module for early action prediction, we conduct extensive experiments on both skeletal and RGB datasets.
We experiment on the NTU RGB+D 60 (NTU60) \cite{shahroudy2016ntu}, NTU RGB+D 120 (NTU120) \cite{shahroudy2016ntu} and SYSU \cite{hu2015jointly} datasets for skeletal data, and the UCF-101 (UCF101) dataset \cite{soomro2012ucf101} for RGB data. 
% Additionally, we also evaluate our proposed ERA module for action recognition task on NTU60 and NTU120. 
% These experiments, along with several other ablation studies, are provided in the Supplementary.
% Besides, we also evaluate our ERA module for action recognition task on NTU60 and NTU120 (with results provided in Supplementary). 

%First Person Hand Action dataset \cite{garcia2018first}, 

%We also obtain state-of-the-art results on UCF101, as observed in Table \ref{tab:ucf_results_pred}.

% To show that our ERA module is flexible enough to be provide improvements on other types of inputs other than skeletal data, we also briefly test our ERA module on the UCF-101 dataset \cite{soomro2012ucf101}. 
% UCF-101 is a popular dataset containing 13,320 video clips of 101 classes of human activities, and is a commonly used dataset for the action prediction  task for RGB video input data. We also obtain state-of-the-art results on UCF-101, as observed in Table \ref{tab:ucf_results_pred}.

% Note that in the 2s-AGCN backbone where graph convolutional modules are used, the ERA module can still be inserted to replace the convolutional kernel within the graph convolutional module. This is because (in the simplest case) the graph convolutional module is just a convolutional module that has additional parameters and steps to account for adjacency information.

% \vspace{-0.17cm}
\subsection{Implementation Details}
% \vspace{-0.1cm}
\textbf{Network Architecture.}
% For the experiments on NTU RGB+D 60/120, SYSU and UCF101, 
% \lgg{other types of params}
Following the previous works \cite{li2020hardnet,wang2019progressive}, we use 2s-AGCN \cite{shi2019two} and 3D ResNeXt-101 \cite{hara2018can} as the backbone networks for skeletal and RGB datasets, respectively. 
% Note that in the 2s-AGCN backbone where graph convolutional modules are used, the ERA module can still be inserted to replace the convolutional kernel within the graph convolutional module. This is because (in the simplest case) the graph convolutional module is just a convolutional module that has additional parameters and steps to account for adjacency information. 
As mentioned above, since the ERA module serves as a plug-and-play replacement for the conventional convolutional module, we uniformly replace $25\%$ of convolutional layers with our ERA module in the backbone networks.
% \lgg{insert into supplementary more details?}
Network hyperparameters $N_{in},N_{out},b_h,b_w$ at each layer follow the original settings in the backbone networks.
% Also, in each ERA module, $d = 0.2 N_{out}$, i.e. $80\%$ of the convolutional kernels are non-expert kernels and the other $20\%$ are expert kernels, while $M=5$ (allowing $5^d$ total expert combinations for each ERA module). 
Also, in each ERA module, $d = 0.2 N_{out}$, i.e. $80\%$ of the convolutional kernels are non-expert kernels and the other $20\%$ are expert kernels, and $M=5$. 
% Compared to the backbone network, our ERA-Net only introduces approximately 20\% additional parameters yet achieves significant performance gain.

For the mapping function $f^{p}$ in Eq.~\ref{eqn:query_mapping}, we first conduct average pooling across the spatial and temporal dimensions of the feature map before a linear layer is used to downsample to the dimensionality $K$, where $K$ is set to $64$.
% \lgg{We first conduct average pooling over the channels, then a fully connected layer is used.}

% We find that pooling functions can be an effective and efficient choice for mapping function $f$ used in Eqn.~\ref{eqn:query_mapping} that is simple and does not add additional parameters. \lgg{We first conduct average pooling over the channels, then a fully connected layer is used.}
% \lgg{We first conduct average pooling over the channels, then max pooling is done spatially to implement $f$.}
% 
% 

\textbf{Training.} We perform experiments on Nvidia RTX 3090 GPU. For skeletal datasets, NTU60, NTU120 and SYSU, we follow \cite{shi2019two} and set the initial learning rate $\alpha$ as $0.1$, which then gradually decays to $0.001$. The batch size $B$ is 64. For RGB dataset UCF101, we follow the same experimental settings as \cite{wang2019progressive}. 
% Network parameters $\theta$ and expert learning rates $\beta$ are updated using the cross-entropy loss from the classification task. 
Network parameters $\theta$ and expert learning rates $\beta$ are updated using $\mathcal{L}$ defined in Sec.~\ref{sec:loss}.
$\gamma_s$ is set to 0.1. 
% Each $\beta_{i}^{p}$ is restricted to $[0,1]$.
Each $\beta_{i}^{p}$ is initialized to $\alpha$, and constrained to non-negative values.
To allow end-to-end training of the retrieved experts in Eqn.~\ref{eqn:argmax}, Gumbel-Softmax \cite{jang2016categorical} gradients are computed during backpropagation for the Argmax operation, with temperature $\tau$ set to $1$. 
\subsection{Experiments on Early Action Prediction}
% \vspace{-0.12cm}

\textbf{NTU60 dataset} \cite{shahroudy2016ntu} has been widely used for 3D action recognition and early action prediction. It is a large dataset that contains more than 56 thousand skeletal sequences from 60 activity classes. All human skeletons in the dataset contain 3D coordinates of 25 body joints. As noted in \cite{li2020hardnet}, this dataset is challenging for the 3D early action prediction task due to the presence of many classes with very similar starting sequences. We follow the evaluation protocol of \cite{li2020hardnet}.
%Two standard evaluation protocols are tested for NTU60. In the Cross Subject (xsub) protocol, half of the subjects are employed for training and the remaining subjects are left for testing. In the Cross View (xview) protocol, two viewpoints are employed for training, and the third one is for testing. 
% \lgg{We use insert as the baseline model}

\begin{table}[t]
\scriptsize
% \tiny
\caption{Performance comparison (\%) of Early Action Prediction on NTU60 and SYSU.
% \lgg{merge NTU60 and SYSU}
We follow the evaluation setting of \cite{li2020hardnet,weng2020early,pang2019dbdnet} and \cite{wang2019progressive} respectively. 
Even without ELRO, we can attain state-of-the-art performance. With ELRO, our method obtains further improvements.
}
% \vspace{-0.3cm}
\centering
% \begin{tabular}{l p{0.4cm} p{0.4cm} p{0.4cm} p{0.4cm} p{0.4cm} p{0.4cm}} 
\begin{tabular}{l | c c c c c c | c c c c c c} 
\toprule
\multirow{2}{*}{Methods} & \multicolumn{6}{c}{Observation Ratios on NTU60} &  \multicolumn{6}{|c}{Observation Ratios on SYSU} \\
\cline{2-7} \cline{8-13}
& 20\% & 40\% & 60\% & 80\% & 100\% & AUC  & 20\% & 40\% & 60\% & 80\% & 100\% & AUC \\
\midrule 
Jain \textit{et al.} \cite{jain2016recurrent} & 7.07 & 18.98 & 44.55 & 63.84 & 71.09 & 37.38  & 31.61 & 53.37 & 68.71 & 73.96 & 75.53 & 57.23 \\
Ke \textit{et al.} \cite{ke2017new} & 8.34 & 26.97 & 56.78 & 75.13 & 80.43 & 45.63  & 26.76 & 52.86 & 72.32 & 79.40 & 80.71 & 58.89  \\
Kong \textit{et al.} \cite{kong2017deep} & - & - & - & - & - & - & 51.75 & 58.83 & 67.17 & 73.83 & 74.67 & 61.33  \\
Ma \textit{et al.} \cite{ma2016learning} & - & - & - & - & - & - & 57.08 & 71.25 & 75.42 & 77.50 & 76.67 & 67.85  \\
Weng \textit{et al.} \cite{weng2020early} & 35.56 & 54.63 & 67.08 & 72.91 & 75.53 & 57.51  & - & - & - & - & - & - \\
\tiny{Aliakbarian \textit{et al.}}\cite{sadegh2017encouraging} & 27.41 & 59.26 & 72.43 & 78.10 & 79.09 & 59.98 & 56.11 & 71.01 & 78.39 & 80.31 & 78.50 & 69.12  \\
Hu \textit{et al.} \cite{hu2018early} & - & - & - & - & - & -  & 56.67 & 75.42 & 80.42 & 82.50 & 79.58 & 71.25  \\
Wang \textit{et al.} \cite{wang2019progressive} & 35.85 & 58.45 & 73.86 & 80.06 & 82.01 & 60.97 & 63.33 & 75.00 & 81.67 & 86.25 & 87.92 & 74.31  \\
Pang \textit{et al.} \cite{pang2019dbdnet} & 33.30 & 56.94 & 74.50 & 80.51 & 81.54 & 61.07 & - & - & - & - & - & - \\
Tran \textit{et al.} \cite{tran2021progressive} & 24.60 & 57.70 & 76.90 & 85.70 & 88.10 & 62.80 & - & - & - & - & - & - \\
Ke \textit{et al.} \cite{ke2019learning} & 32.12 & 63.82 & 77.02 & 82.45 &  83.19 & 64.22 & 58.81 & 74.21 & 82.18 & 84.42 & 83.14 & 72.55  \\
HARD-Net \cite{li2020hardnet} & 42.39 & 72.24 & 82.99 & 86.75 & 87.54 & 70.56 & - & - & - & - & - & -   \\
\midrule
Baseline & 38.09 & 66.36 & 78.67 & 83.29 & 84.10 & 66.43   & 60.71 & 73.04 & 77.81 & 83.88 & 84.32 & 72.20  \\
\tiny{ERA-Net w/o ELRO} & 43.94 & 73.23 & 84.53 & 87.61 & 87.97 & 71.62  & 63.50 & 80.82 & 82.70 & 86.33 & 87.10 & 75.78 \\
ERA-Net & \textbf{53.98} & \textbf{74.34} & \textbf{85.03} & \textbf{88.35} & \textbf{88.45} & \textbf{73.87}   & \textbf{65.30} & \textbf{81.27} & \textbf{85.67} & \textbf{89.17} & \textbf{89.38} & \textbf{77.73} \\
\bottomrule
\end{tabular}
\label{tab:ntu60_sysu_results_pred} 
% \vspace{-0.6cm}
\end{table}

We first compare the proposed ERA-Net with the state-of-the-art approaches on NTU60. The results over different observation ratios are shown in Table \ref{tab:ntu60_sysu_results_pred}.
Our full method is employed in the \textbf{ERA-Net} setting.  
% In \textbf{ERA-Net w/o ELRO}, we use ERA modules but do not implement $\beta$ to train experts using the proposed Expert Learning Rate Optimization algorithm, instead we train using a single backpropagation step that updates all 
In \textbf{ERA-Net w/o ELRO}, we use ERA modules but do not implement $\beta$ to train experts using the proposed ELRO algorithm, instead we train using a single backpropagation step that updates all 
model parameters at each iteration.  
We also provide the \textbf{Baseline} setting for comparison, where the backbone is used without ERA modules.

We report the prediction accuracy at each observation ratio. Furthermore,
we use the
Area Under Curve (AUC) metric in our experiments, following previous works \cite{li2020hardnet,weng2020early,pang2019dbdnet}. The AUC measures the average precision over all observation ratios and broadly summarizes each model's performance into a single metric. On NTU60, we achieve more than a 3 point improvement against existing state-of-the art method \cite{li2020hardnet}, suggesting that the ERA module effectively increases the discriminative capabilities on the early action prediction task.  %\lgg{insert how much we outperform the other models, current sota.}  %, a 7 point improvement on NTU120, more than 3 point improvement on both SYSU and UCF101.

One crucial observation is that ERA-Net outperforms existing methods more significantly when the observation ratio is low. For example, when the observation ratio is $20\%$, ERA-Net improves over state-of-the-art \cite{li2020hardnet} by more than $11\%$, %, but when the observation ratio increases to $100\%$, we only improve by roughly $1\%$.  As the ERA module improves on baseline methods more when the setting is more challenging, this 
which further demonstrates that the ERA module is especially effective in picking up subtle cues to tackle hard samples (where samples are more similar at the earlier stages).

% \subsection{Early Action Prediction on SYSU}
\noindent\textbf{SYSU dataset} \cite{hu2015jointly} is also commonly used for 3D action recognition and early action prediction. The dataset contains 480 skeletal sequences belonging to 12 action classes performed by 40 subjects. The human skeletons in this dataset contain 3D coordinates of 20 joints. We follow evaluation protocol of \cite{wang2019progressive}. Comparisons against state-of-the-art methods are displayed in Table \ref{tab:ntu60_sysu_results_pred},
where ERA-Net outperforms the current state-of-the-art \cite{wang2019progressive} by about 3 points.

% %%original wrapped table
% \begin{wraptable}[9]{r}{0.6\linewidth}
% \scriptsize
% \caption{Performance comparison (\%) of Early Action Prediction on NTU120.
% Since no previous works report early action prediction results on NTU120, we compare our method to the baseline.
% }
% \centering
% % \begin{tabular}{l p{0.4cm} p{0.4cm} p{0.4cm} p{0.4cm} p{0.4cm} p{0.4cm}} 
% \begin{tabular}{l c c c c c c} 
% \toprule
% \multirow{2}{*}{Methods} & \multicolumn{6}{c}{Observation Ratios} \\
% \cline{2-7}
% & 20\% & 40\% & 60\% & 80\% & 100\% & AUC \\
% \midrule 
% Baseline & 23.14 & 32.49 & 59.07 & 75.61 & 81.18 & 50.03 \\
% ERA-Net w/o ELRO & 29.60 & 43.45 & 65.14 & 78.03 & 82.01 & 55.52\\
% ERA-Net & \textbf{31.73} & \textbf{45.67} & \textbf{67.08} & \textbf{78.84} & \textbf{82.43} & \textbf{57.02}   \\
% \bottomrule
% \end{tabular}
% \label{tab:ntu_120_results_pred} 
% \end{wraptable}

% Since no previous works report early action prediction results on NTU120, we compare our method to the baseline.
% We follow the evaluation setting of \cite{wang2019progressive}.
\begin{table}[t]
\scriptsize
% \tiny
\caption{Performance comparison (\%) of Early Action Prediction on NTU120 and UCF101. 
As no prior works report NTU120 early action prediction results, we compare our method to the baseline.
For UCF101, we follow the evaluation setting of \cite{wang2019progressive}.
% For UCF101, we follow the evaluation setting of \cite{wang2019progressive}.
% As no previous works report early action prediction results on NTU120, we compare our method to the baseline.
% Even without ELRO, we can attain state-of-the-art performance. With ELRO, our method obtains further improvements.
}
% \vspace{-0.3cm}
\centering
% \begin{tabular}{l p{0.4cm} p{0.4cm} p{0.4cm} p{0.4cm} p{0.4cm} p{0.4cm}} 
\begin{tabular}{l | c c c c c c | c c c c c c} 
\toprule
\multirow{2}{*}{Methods} & \multicolumn{6}{c}{Observation Ratios on NTU120} &  \multicolumn{6}{|c}{Observation Ratios on UCF101} \\
\cline{2-7} \cline{8-13}
& 20\% & 40\% & 60\% & 80\% & 100\% & AUC  & 10\% & 30\% & 50\% & 70\% & 90\% & AUC \\
\midrule 
MSRNN \cite{hu2018early} & - & - & - & - & - & -  & 68.01 & 88.71 & 89.25 & 89.92 & 90.23 & 80.89 \\
Wu \textit{et al.} \cite{wu2021spatial} & - & - & - & - & - & - & 80.24 & 84.55 & 86.28 & 87.53 & 88.24 & 80.57 \\
Wu \textit{et al.} \cite{wu2021anticipating} & - & - & - & - & - & -  & 82.36 & 88.97 & 91.32 & 92.41 & 93.02 & 84.66 \\
Wang \textit{et al.} \cite{wang2019progressive} & - & - & - & - & - & -  & 83.32 & 88.92 & 90.85 & 91.28 & 91.31 & 89.64  \\
\midrule
Baseline & 23.14 & 32.49 & 59.07 & 75.61 & 81.18 & 50.03 &  82.88 & 89.02 & 89.64 & 91.12 & 91.96 & 89.30 \\
\tiny{ERA-Net w/o ELRO} & 29.60 & 43.45 & 65.14 & 78.03 & 82.01 & 55.52 &  86.99 & 91.49 & 93.63 & 94.24 & 94.40 & 92.51 \\
ERA-Net & \textbf{31.73} & \textbf{45.67} & \textbf{67.08} & \textbf{78.84} & \textbf{82.43} & \textbf{57.02}  &  \textbf{89.14} & \textbf{92.39} & \textbf{94.29} & \textbf{95.45} & \textbf{95.72} & \textbf{93.64}  \\
\bottomrule
\end{tabular}
\label{tab:ntu120_ucf101_results_pred} 
% \vspace{-0.05cm}
\end{table}

% \subsection{Early Action Prediction on NTU120}
\noindent\textbf{NTU120 dataset} \cite{liu2019ntu} is an extension of NTU60. It is currently the largest RGB+D dataset for 3D action analysis with more than 114k skeletal sequences and contains 120 activity classes. This dataset is challenging for the early action prediction task, containing many classes that are hard to classify without observing the full sequences. Comparisons are displayed in Table \ref{tab:ntu120_ucf101_results_pred},
where ERA-Net outperforms the baseline by about 7 points on the AUC metric. We also observe very large improvements at lower observation ratios,
demonstrating the efficacy of our method for early action prediction.

%%%original wrapped table
% \begin{wraptable}[11]{r}{0.625\linewidth}
% \vspace{-0.3cm}
% \scriptsize
% \caption{Performance comparison (\%) of Early Action Prediction on UCF101. We follow the evaluation setting of \cite{wang2019progressive}.
% % \vspace{-0.1cm}
% }
% \centering
% % \begin{tabular}{l p{0.4cm} p{0.4cm} p{0.4cm} p{0.4cm} p{0.4cm} p{0.4cm}} 
% \begin{tabular}{l c c c c c c} 
% \toprule
% \multirow{2}{*}{Methods} & \multicolumn{6}{c}{Observation Ratios} \\
% \cline{2-7}
% & 10\% & 30\% & 50\% & 70\% & 90\% & AUC \\
% \midrule 
% % Mem-LSTM \cite{kong2018action} & 51.02 & 86.75 & 88.37 & 89.22 & 89.97 & 77.24 \\
% MSRNN \cite{hu2018early} & 68.01 & 88.71 & 89.25 & 89.92 & 90.23 & 80.89 \\
% Wang \textit{et al.} \cite{wang2019progressive} & 83.32 & 88.92 & 90.85 & 91.28 & 91.31 & 89.64  \\
% Wu \textit{et al.} \cite{wu2021spatial} & 80.24 & 84.55 & 86.28 & 87.53 & 88.24 & 80.57 \\
% Wu \textit{et al.} \cite{wu2021anticipating} & 82.36 & 88.97 & 91.32 & 92.41 & 93.02 & 84.66 \\
% \midrule
% Baseline & 82.88 & 89.02 & 89.64 & 91.12 & 91.96 & 89.30 \\
% ERA-Net w/o ELRO & 86.99 & 91.49 & 93.63 & 94.24 & 94.40 & 92.51 \\
% ERA-Net & \textbf{89.14} & \textbf{92.39} & \textbf{94.29} & \textbf{95.45} & \textbf{95.72} & \textbf{93.64}  \\
% \bottomrule
% \end{tabular}
% \label{tab:ucf_results_pred} 
% \end{wraptable}

% \subsection{Early Action Prediction on UCF101}
\noindent\textbf{UCF101 dataset} \cite{soomro2012ucf101} is a popular dataset containing 13,320 video clips of 101 classes of human activities. It is a commonly used dataset for action prediction from RGB videos. 
% \lgg{We use insert as the baseline model} 
Comparisons against state-of-the-art action prediction methods are shown in Table \ref{tab:ntu120_ucf101_results_pred}, where ERA-Net outperforms current state-of-the-art methods \cite{wu2021anticipating,wu2021spatial,wang2019progressive} by 4 or more AUC points,  
% \textcolor{red}{3 points? other baselines??? you need to be specific, what metric and it outperforms which baseline or SOTA model?}, 
showing that ERA provides gains for early action prediction on RGB video datasets as well.

\subsection{Ablation Study}
\label{section:ablation}

\noindent{\textbf{Impact of number of experts.}}
We evaluate the ratio of experts and non-experts in Table~\ref{tab:ablation_all}\textcolor{red}{(a)}. As performance peaks at $20:80$, we set $d = 0.2 N_{out}$,
% \lgg{Above this ratio, the benefit of increasing discriminativeness on subtle cues might be outweighed by the reduced encoding of general patterns. (to edit, might change)}
which allows for encoding of the most effective mix of general patterns and subtle cues within the layer. 

\begin{table}[t]
\scriptsize
\caption{
Ablation studies conducted on NTU60. 
(a) Evaluation of ratios between number of experts and non-experts;
(b) evaluation of size of Expert Banks $M$;
(c) evaluation of the percentage (\%) of convolutional layers replaced by ERA modules;
(d) evaluation of the value of similarity loss weight $\gamma_s$;
(e) evaluation of our dynamic retrieval mechanism against alternative static designs.
% \lgg{edit caption. merging all tables here...}
% \vspace{-0.2cm}
}
\centering
% \begin{tabular}{||c | c || c | c || c | c || c | c ||} 
\begin{tabular}{||c | c || c | c || c | c || c | c || c | c ||} 
\hline
\multicolumn{2}{||c||}{(a)} & \multicolumn{2}{|c||}{(b)} & \multicolumn{2}{|c||}{(c)} & \multicolumn{2}{|c||}{(d)} & \multicolumn{2}{|c||}{(e)}  \\
\hline
% Expert:Non-expert & AUC & M & AUC & \% of ERA modules & AUC & $\gamma_s$ & AUC \\ 
Expert:Non-expert & AUC & $M$ & AUC & \% of ERA modules & AUC & $\gamma_s$ & AUC & Method & AUC \\ 
\hline
0:100 & 66.43 & 1 & 66.43 & 0 & 66.43 & 0.05 & 72.92 & Extra-Channel & 67.55 \\
20:80 & 73.87 & 2 & 71.55 & 25 & 73.87 & 0.1 & 73.87 & Expert-Avg  & 68.02 \\
60:40 & 71.22 & 5 & 73.87 & 50 & 73.79 & 0.2 & 73.85 & ERA-Net  & 73.87 \\
100:0 & 70.12 & 10 & 73.86 & 100 & 73.81 & 0.3 & 73.76  &  &  \\
    %   &       & 10 & 73.88 &   &  &  &  \\
\hline
\end{tabular}
% \label{tab:ablation_number_experts} 
\label{tab:ablation_all} 
% \vspace{-0.2cm}
\end{table}
%%% take out one M value 3 & 73.87

% %%%%%%%%% original submission %%%%%%%%%%%
% \noindent{\textbf{Impact of size of Expert Banks ($M$).}} We evaluate the size of Expert Banks in Table~\ref{tab:ablation_all}\textcolor{red}{(b)}. 
% We find that the performance increases moderately when $M$ is increased from $2$ to $5$, and remains stable when we further increase it. 
% We argue {% tex broke the page here!!!!
% \parfillskip=0pt
% \parskip=0pt
% \par}
% \begin{wrapfigure}{r}{0.7\linewidth}
%     \centering
%     % \vspace{-2cm}
%     \includegraphics[trim={0 0 0 0}, width=\linewidth]{figs/visulization_matching_score_v5.pdf}
%     % \vspace{-0.8cm}
%     \caption{
%     Visualization of experts selection at 20\% observation of the actions. 
%     Each subplot contains information from an Expert Bank, where the horizontal axis represents the $M$ experts (where $M=5$) and the vertical axis denotes their normalized matching scores.
%     (Left) At 20\% observation ratio, actions ``Punch/Slap'' and ``Point finger" are similar, and their expert matching scores are also similar.
%     (Right) At 20\% observation ratio, actions ``Point finger'' and ``Kicking" are different, and their expert matching scores are also different.    
%     }
%     \label{fig:expert_selection_visualization}
% % \vspace{-0.3cm}
% \end{wrapfigure}
% \noindent that this is because 
% the representation capacity by setting $M=5$ is sufficient to capture the subtle cues present in the dataset. 

%%%%%%%%% camera ready version %%%%%%%%%%%
\noindent{\textbf{Impact of size of Expert Banks ($M$).}} We evaluate the size of Expert Banks in Table~\ref{tab:ablation_all}\textcolor{red}{(b)}. 
We find that the performance increases moderately when $M$ is increased from $2$ to $5$, and remains stable when we further increase it. 
We argue that this is because 
the representation capacity by setting $M=5$ is sufficient to capture the subtle cues present in the dataset.

\noindent{\textbf{Impact of number of ERA modules.}}
We ablate the decision of replacing $25\%$ of convolutional layers with ERA modules in Table~\ref{tab:ablation_all}\textcolor{red}{(c)}. We find that, above $25\%$, the performance does not increase further.  %\textcolor{green}{again we need to explain why. }
% \lgg{This might be because the representation capacity is already enough to tackle the subtle cues in the dataset at 25\%}
This suggests that, at 25\%, there is already sufficient representation capacity to handle the encoding of subtle cues.

\noindent \textbf{Impact of similarity loss weight ($\gamma_s$).} We conduct ablation studies on the impact of $\gamma_s$ in Table \ref{tab:ablation_all}\textcolor{red}{(d)}. $\gamma_s = 0.1$ performs the best.
% \lgg{Further visualization depicted in Fig x shows that, 
% \lgg{Further visualization shows that, when $\gamma_s$ is set too high,}
% This possibly because, when $\gamma_s$ is set too low, the experts are not as diverse, and when it is set too high, the experts are more diverse (but may not perform as well).
% (but may not perform as well)
% This possibly because \textcolor{green}{ can we delete "possibly"?? why we are not sure?}, 
This because, when $\gamma_s$ is set too low, the experts are not as diverse, and when it is set too high, the experts may lose focus on the main objective.

% \begin{table}[ht]
% \footnotesize
% \caption{
% % Latency (ms) incurred during testing. 
% Evaluations of our dynamic retrieval mechanism against alternative static designs on NTU60. 
% }
% \centering
% \begin{tabular}{l c } 
% \toprule
% Method & AUC \\ 
% \midrule
% Extra-Channel & 67.55 \\
% Expert-Avg  & 68.02 \\
% ERA-Net  & 73.87  \\
% \bottomrule
% \end{tabular}
% \label{tab:dynamic_static} 
% \end{table}

% \begin{wraptable}{r}{4cm}
% \footnotesize
% \caption{
% % Latency (ms) incurred during testing. 
% Evaluations of our dynamic retrieval mechanism against alternative static designs on NTU60. 
% }
% \centering
% \begin{tabular}{l c } 
% \toprule
% Method & AUC \\ 
% \midrule
% Extra-Channel & 67.55 \\
% Expert-Avg  & 68.02 \\
% ERA-Net  & 73.87  \\
% \bottomrule
% \end{tabular}
% \label{tab:dynamic_static} 
% \vspace{-1.5cm}
% \end{wraptable}

\noindent \textbf{Impact of dynamic retrieval mechanism.}
We evaluate our dynamic design by comparing our ERA module against other alternative static designs in Table \ref{tab:ablation_all}\textcolor{red}{(e)}.
% \textbf{Expert-Avg} averages the outputs of all experts within the Expert Bank, 
\textbf{Expert-Avg} averages the outputs of all experts within the Expert Bank (i.e. all experts are  used for each input sample, without dynamic expert selection),
while \textbf{Extra-Channel} adds extra channels to the traditional convolutional layer.
\textit{Notably, these alternative static designs use the same number of parameters as our ERA-Net.}
We find that our dynamic retrieval mechanism provides significant improvement over these alternatives.
% \lgg{An experiment to account for the extra parameters???? Vs average of all experts. Or just add those extra channels to the conv layer.}

% \textbf{Impact of number of experts $M$.} \lgg{tuning of hyperparameters on multiple datasets (some parts already in Supp)}

% \lgg{An experiment on fine-grained action recognition?? (Supp)}

% \lgg{Supp: tune the penalty parameter.}

% \lgg{vs conditional convolution. our fixed model, not growing.}

% \lgg{add experimental details into the supplementary material. especially UCF101}

% \noindent \textbf{Fine-grained action recognition on Diving48 dataset.} In principle, our ERA module will also provide improvements on other tasks that involve subtle differences. Thus, we evaluate our ERA module on fine-grained action recognition, using a widely used Diving48 dataset \cite{li2018diving48}. \lgg{Using 3D ResNext101 as a baseline, we observe improvements of x\% (y\% vs z\%) when our ERA module is used.}
% % \lgg{Results are shown in Table x, where we provide significant improvements upon the baseline 3d ResNext101.}

% \vspace{-0.2cm}
\section{Conclusion}
% \vspace{-0.2cm}
In this paper, we have proposed a novel plug-and-play ERA module for early action prediction. 
To encourage the experts to effectively use subtle differences for early action prediction, we push them to discriminate exclusively among similar samples.
An Expert Learning Rate Optimization algorithm is further proposed to balance the training among numerous experts, which improves performance. 
Our method obtains state-of-the-art performance on four popular datasets.

% \footnotesize \noindent \textbf{Acknowledgement} This work is supported by National Research Foundation, Singapore under its AI Singapore Programme (AISG Award No: AISG-100E-2020-065), SUTD Startup Research Grant and MOE Tier 1 Grant.

\footnotesize \noindent \textbf{Acknowledgement}
This work is supported by National Research Foundation, Singapore under its AI Singapore Programme (AISG Award No: AISG-100E-2020-065), Ministry of Education Tier 1 Grant and SUTD Startup Research Grant.

\clearpage
\bibliographystyle{splncs04}
\bibliography{camera_ready}

\begin{thebibliography}{10}
\providecommand{\url}[1]{\texttt{#1}}
\providecommand{\urlprefix}{URL }
\providecommand{\doi}[1]{https://doi.org/#1}

\bibitem{chaabane2020looking}
Chaabane, M., Trabelsi, A., Blanchard, N., Beveridge, R.: Looking ahead:
  Anticipating pedestrians crossing with future frames prediction. In:
  Proceedings of the IEEE/CVF Winter Conference on Applications of Computer
  Vision. pp. 2297--2306 (2020)

\bibitem{chen2020recurrent}
Chen, L., Lu, J., Song, Z., Zhou, J.: Recurrent semantic preserving generation
  for action prediction. IEEE Transactions on Circuits and Systems for Video
  Technology  \textbf{31}(1),  231--245 (2020)

\bibitem{chen2020dynamic}
Chen, Y., Dai, X., Liu, M., Chen, D., Yuan, L., Liu, Z.: Dynamic convolution:
  Attention over convolution kernels. In: Proceedings of the IEEE/CVF
  Conference on Computer Vision and Pattern Recognition. pp. 11030--11039
  (2020)

\bibitem{chen2021ctrgcn}
Chen, Y., Zhang, Z., Yuan, C., Li, B., Deng, Y., Hu, W.: Channel-wise topology
  refinement graph convolution for skeleton-based action recognition. In:
  Proceedings of the IEEE/CVF International Conference on Computer Vision. pp.
  13359--13368 (2021)

\bibitem{cheng2020skeleton}
Cheng, K., Zhang, Y., He, X., Chen, W., Cheng, J., Lu, H.: Skeleton-based
  action recognition with shift graph convolutional network. In: Proceedings of
  the IEEE/CVF Conference on Computer Vision and Pattern Recognition. pp.
  183--192 (2020)

\bibitem{emad2021early}
Emad, M., Ishack, M., Ahmed, M., Osama, M., Salah, M., Khoriba, G.:
  Early-anomaly prediction in surveillance cameras for security applications.
  In: 2021 International Mobile, Intelligent, and Ubiquitous Computing
  Conference (MIUCC). pp. 124--128. IEEE (2021)

\bibitem{fatima2013unified}
Fatima, I., Fahim, M., Lee, Y.K., Lee, S.: A unified framework for activity
  recognition-based behavior analysis and action prediction in smart homes.
  Sensors  \textbf{13}(2),  2682--2699 (2013)

\bibitem{feichtenhofer2019slowfast}
Feichtenhofer, C., Fan, H., Malik, J., He, K.: Slowfast networks for video
  recognition. In: Proceedings of the IEEE/CVF international conference on
  computer vision. pp. 6202--6211 (2019)

\bibitem{finn2017maml}
Finn, C., Abbeel, P., Levine, S.: Model-agnostic meta-learning for fast
  adaptation of deep networks. In: International Conference on Machine
  Learning. pp. 1126--1135. PMLR (2017)

\bibitem{gammulle2019predicting}
Gammulle, H., Denman, S., Sridharan, S., Fookes, C.: Predicting the future: A
  jointly learnt model for action anticipation. In: Proceedings of the IEEE/CVF
  International Conference on Computer Vision. pp. 5562--5571 (2019)

\bibitem{gujjar2019classifying}
Gujjar, P., Vaughan, R.: Classifying pedestrian actions in advance using
  predicted video of urban driving scenes. In: 2019 International Conference on
  Robotics and Automation (ICRA). pp. 2097--2103. IEEE (2019)

\bibitem{han2021dynamic}
Han, Y., Huang, G., Song, S., Yang, L., Wang, H., Wang, Y.: Dynamic neural
  networks: A survey. arXiv preprint arXiv:2102.04906  (2021)

\bibitem{hara2018can}
Hara, K., Kataoka, H., Satoh, Y.: Can spatiotemporal 3d cnns retrace the
  history of 2d cnns and imagenet? In: Proceedings of the IEEE conference on
  Computer Vision and Pattern Recognition. pp. 6546--6555 (2018)

\bibitem{hu2015jointly}
Hu, J.F., Zheng, W.S., Lai, J., Zhang, J.: Jointly learning heterogeneous
  features for rgb-d activity recognition. In: Proceedings of the IEEE
  conference on computer vision and pattern recognition. pp. 5344--5352 (2015)

\bibitem{hu2016real}
Hu, J.F., Zheng, W.S., Ma, L., Wang, G., Lai, J.: Real-time rgb-d activity
  prediction by soft regression. In: European Conference on Computer Vision.
  pp. 280--296. Springer (2016)

\bibitem{hu2018early}
Hu, J.F., Zheng, W.S., Ma, L., Wang, G., Lai, J., Zhang, J.: Early action
  prediction by soft regression. IEEE transactions on pattern analysis and
  machine intelligence  \textbf{41}(11),  2568--2583 (2018)

\bibitem{huang2016anticipatory}
Huang, C.M., Mutlu, B.: Anticipatory robot control for efficient human-robot
  collaboration. In: 2016 11th ACM/IEEE international conference on human-robot
  interaction (HRI). pp. 83--90. IEEE (2016)

\bibitem{jain2016recurrent}
Jain, A., Singh, A., Koppula, H.S., Soh, S., Saxena, A.: Recurrent neural
  networks for driver activity anticipation via sensory-fusion architecture.
  In: 2016 IEEE International Conference on Robotics and Automation (ICRA). pp.
  3118--3125. IEEE (2016)

\bibitem{jang2016categorical}
Jang, E., Gu, S., Poole, B.: Categorical reparameterization with
  gumbel-softmax. arXiv preprint arXiv:1611.01144  (2016)

\bibitem{johnson2019survey}
Johnson, J.M., Khoshgoftaar, T.M.: Survey on deep learning with class
  imbalance. Journal of Big Data  \textbf{6}(1),  1--54 (2019)

\bibitem{ke2017new}
Ke, Q., Bennamoun, M., An, S., Sohel, F., Boussaid, F.: A new representation of
  skeleton sequences for 3d action recognition. In: Proceedings of the IEEE
  conference on computer vision and pattern recognition. pp. 3288--3297 (2017)

\bibitem{ke2019learning}
Ke, Q., Bennamoun, M., Rahmani, H., An, S., Sohel, F., Boussaid, F.: Learning
  latent global network for skeleton-based action prediction. IEEE Transactions
  on Image Processing  \textbf{29},  959--970 (2019)

\bibitem{kong2018human}
Kong, Y., Fu, Y.: Human action recognition and prediction: A survey. arXiv
  preprint arXiv:1806.11230  (2018)

\bibitem{kong2018action}
Kong, Y., Gao, S., Sun, B., Fu, Y.: Action prediction from videos via
  memorizing hard-to-predict samples. In: Proceedings of the AAAI Conference on
  Artificial Intelligence. vol.~32 (2018)

\bibitem{kong2014discriminative}
Kong, Y., Kit, D., Fu, Y.: A discriminative model with multiple temporal scales
  for action prediction. In: European conference on computer vision. pp.
  596--611. Springer (2014)

\bibitem{kong2017deep}
Kong, Y., Tao, Z., Fu, Y.: Deep sequential context networks for action
  prediction. In: Proceedings of the IEEE conference on computer vision and
  pattern recognition. pp. 1473--1481 (2017)

\bibitem{kong2018adversarial}
Kong, Y., Tao, Z., Fu, Y.: Adversarial action prediction networks. IEEE
  transactions on pattern analysis and machine intelligence  \textbf{42}(3),
  539--553 (2018)

\bibitem{koppula2015anticipating}
Koppula, H.S., Saxena, A.: Anticipating human activities using object
  affordances for reactive robotic response. IEEE transactions on pattern
  analysis and machine intelligence  \textbf{38}(1),  14--29 (2015)

\bibitem{li20212d}
Li, H., Wu, Z., Shrivastava, A., Davis, L.S.: 2d or not 2d? adaptive 3d
  convolution selection for efficient video recognition. In: CVPR. pp.
  6155--6164 (2021)

\bibitem{li2020hardnet}
Li, T., Liu, J., Zhang, W., Duan, L.: Hard-net: Hardness-aware discrimination
  network for 3d early activity prediction. In: European Conference on Computer
  Vision. pp. 420--436. Springer (2020)

\bibitem{lin2019tsm}
Lin, J., Gan, C., Han, S.: Tsm: Temporal shift module for efficient video
  understanding. In: Proceedings of the IEEE/CVF International Conference on
  Computer Vision. pp. 7083--7093 (2019)

\bibitem{liu2019ntu}
Liu, J., Shahroudy, A., Perez, M., Wang, G., Duan, L.Y., Kot, A.C.: Ntu rgb+ d
  120: A large-scale benchmark for 3d human activity understanding. IEEE
  transactions on pattern analysis and machine intelligence  \textbf{42}(10),
  2684--2701 (2019)

\bibitem{liu2019skeleton}
Liu, J., Shahroudy, A., Wang, G., Duan, L.Y., Kot, A.C.: Skeleton-based online
  action prediction using scale selection network. IEEE transactions on pattern
  analysis and machine intelligence  \textbf{42}(6),  1453--1467 (2019)

\bibitem{liu2020disentangling}
Liu, Z., Zhang, H., Chen, Z., Wang, Z., Ouyang, W.: Disentangling and unifying
  graph convolutions for skeleton-based action recognition. In: Proceedings of
  the IEEE/CVF conference on computer vision and pattern recognition. pp.
  143--152 (2020)

\bibitem{ma2016learning}
Ma, S., Sigal, L., Sclaroff, S.: Learning activity progression in lstms for
  activity detection and early detection. In: Proceedings of the IEEE
  conference on computer vision and pattern recognition. pp. 1942--1950 (2016)

\bibitem{mavrogiannis2020b}
Mavrogiannis, A., Chandra, R., Manocha, D.: B-gap: Behavior-guided action
  prediction for autonomous navigation. arXiv preprint arXiv:2011.03748  (2020)

\bibitem{mullapudi2018hydranets}
Mullapudi, R.T., Mark, W.R., Shazeer, N., Fatahalian, K.: Hydranets:
  Specialized dynamic architectures for efficient inference. In: Proceedings of
  the IEEE Conference on Computer Vision and Pattern Recognition. pp.
  8080--8089 (2018)

\bibitem{nguyen2021geomnet}
Nguyen, X.S.: Geomnet: A neural network based on riemannian geometries of spd
  matrix space and cholesky space for 3d skeleton-based interaction
  recognition. In: Proceedings of the IEEE/CVF International Conference on
  Computer Vision. pp. 13379--13389 (2021)

\bibitem{pang2019dbdnet}
Pang, G., Wang, X., Hu, J., Zhang, Q., Zheng, W.S.: Dbdnet: Learning
  bi-directional dynamics for early action prediction. In: IJCAI. pp. 897--903
  (2019)

\bibitem{reily2018skeleton}
Reily, B., Han, F., Parker, L.E., Zhang, H.: Skeleton-based bio-inspired human
  activity prediction for real-time human--robot interaction. Autonomous Robots
   \textbf{42}(6),  1281--1298 (2018)

\bibitem{sadegh2017encouraging}
Sadegh~Aliakbarian, M., Sadat~Saleh, F., Salzmann, M., Fernando, B., Petersson,
  L., Andersson, L.: Encouraging lstms to anticipate actions very early. In:
  Proceedings of the IEEE International Conference on Computer Vision. pp.
  280--289 (2017)

\bibitem{shahroudy2016ntu}
Shahroudy, A., Liu, J., Ng, T.T., Wang, G.: Ntu rgb+ d: A large scale dataset
  for 3d human activity analysis. In: Proceedings of the IEEE conference on
  computer vision and pattern recognition. pp. 1010--1019 (2016)

\bibitem{shazeer2017outrageously}
Shazeer, N., Mirhoseini, A., Maziarz, K., Davis, A., Le, Q., Hinton, G., Dean,
  J.: Outrageously large neural networks: The sparsely-gated mixture-of-experts
  layer (2017)

\bibitem{shi2019skeleton}
Shi, L., Zhang, Y., Cheng, J., Lu, H.: Skeleton-based action recognition with
  directed graph neural networks. In: Proceedings of the IEEE/CVF Conference on
  Computer Vision and Pattern Recognition. pp. 7912--7921 (2019)

\bibitem{shi2019two}
Shi, L., Zhang, Y., Cheng, J., Lu, H.: Two-stream adaptive graph convolutional
  networks for skeleton-based action recognition. In: Proceedings of the
  IEEE/CVF conference on computer vision and pattern recognition. pp.
  12026--12035 (2019)

\bibitem{shi2021adasgn}
Shi, L., Zhang, Y., Cheng, J., Lu, H.: Adasgn: Adapting joint number and model
  size for efficient skeleton-based action recognition. In: Proceedings of the
  IEEE/CVF International Conference on Computer Vision. pp. 13413--13422 (2021)

\bibitem{shu2019metaweightnet}
Shu, J., Xie, Q., Yi, L., Zhao, Q., Zhou, S., Xu, Z., Meng, D.:
  Meta-weight-net: Learning an explicit mapping for sample weighting (2019)

\bibitem{song2020stronger}
Song, Y.F., Zhang, Z., Shan, C., Wang, L.: Stronger, faster and more
  explainable: A graph convolutional baseline for skeleton-based action
  recognition. In: Proceedings of the 28th ACM International Conference on
  Multimedia. pp. 1625--1633 (2020)

\bibitem{soomro2012ucf101}
Soomro, K., Zamir, A.R., Shah, M.: Ucf101: A dataset of 101 human actions
  classes from videos in the wild. arXiv preprint arXiv:1212.0402  (2012)

\bibitem{tran2021progressive}
Tran, V., Balasubramanian, N., Hoai, M.: Progressive knowledge distillation for
  early action recognition. In: 2021 IEEE International Conference on Image
  Processing (ICIP). pp. 2583--2587. IEEE (2021)

\bibitem{veit2018convaig}
Veit, A., Belongie, S.: Convolutional networks with adaptive inference graphs.
  In: Proceedings of the European Conference on Computer Vision (ECCV). pp.
  3--18 (2018)

\bibitem{wang2021ga}
Wang, W., Chang, F., Liu, C., Li, G., Wang, B.: Ga-net: A guidance aware
  network for skeleton-based early activity recognition. IEEE Transactions on
  Multimedia  (2021)

\bibitem{wang2018skipnet}
Wang, X., Yu, F., Dou, Z.Y., Darrell, T., Gonzalez, J.E.: Skipnet: Learning
  dynamic routing in convolutional networks. In: Proceedings of the European
  Conference on Computer Vision (ECCV). pp. 409--424 (2018)

\bibitem{wang2019progressive}
Wang, X., Hu, J.F., Lai, J.H., Zhang, J., Zheng, W.S.: Progressive
  teacher-student learning for early action prediction. In: Proceedings of the
  IEEE/CVF Conference on Computer Vision and Pattern Recognition. pp.
  3556--3565 (2019)

\bibitem{weng2020early}
Weng, J., Jiang, X., Zheng, W.L., Yuan, J.: Early action recognition with
  category exclusion using policy-based reinforcement learning. IEEE
  Transactions on Circuits and Systems for Video Technology  \textbf{30}(12),
  4626--4638 (2020)

\bibitem{wu2021spatial}
Wu, X., Wang, R., Hou, J., Lin, H., Luo, J.: Spatial--temporal relation
  reasoning for action prediction in videos. International Journal of Computer
  Vision  \textbf{129}(5),  1484--1505 (2021)

\bibitem{wu2021anticipating}
Wu, X., Zhao, J., Wang, R.: Anticipating future relations via graph growing for
  action prediction. In: Proceedings of the AAAI Conference on Artificial
  Intelligence. vol.~35, pp. 2952--2960 (2021)

\bibitem{wu2021coarse}
Wu, Z., Li, H., Zheng, Y., Xiong, C., Jiang, Y., Davis, L.S.: A coarse-to-fine
  framework for resource efficient video recognition. IJCV  \textbf{129}(11),
  2965--2977 (2021)

\bibitem{wu2018blockdrop}
Wu, Z., Nagarajan, T., Kumar, A., Rennie, S., Davis, L.S., Grauman, K., Feris,
  R.: Blockdrop: Dynamic inference paths in residual networks. In: CVPR. pp.
  8817--8826 (2018)

\bibitem{xie2018s3d}
Xie, S., Sun, C., Huang, J., Tu, Z., Murphy, K.: Rethinking spatiotemporal
  feature learning: Speed-accuracy trade-offs in video classification. In:
  Proceedings of the European conference on computer vision (ECCV). pp.
  305--321 (2018)

\bibitem{xie2020spatially}
Xie, Z., Zhang, Z., Zhu, X., Huang, G., Lin, S.: Spatially adaptive inference
  with stochastic feature sampling and interpolation. In: European Conference
  on Computer Vision. pp. 531--548. Springer (2020)

\bibitem{xu2019prediction}
Xu, W., Yu, J., Miao, Z., Wan, L., Ji, Q.: Prediction-cgan: Human action
  prediction with conditional generative adversarial networks. In: Proceedings
  of the 27th ACM International Conference on Multimedia. pp. 611--619 (2019)

\bibitem{yan2018participation}
Yan, R., Tang, J., Shu, X., Li, Z., Tian, Q.: Participation-contributed
  temporal dynamic model for group activity recognition. In: Proceedings of the
  26th ACM international conference on Multimedia. pp. 1292--1300 (2018)

\bibitem{yan2020higcin}
Yan, R., Xie, L., Tang, J., Shu, X., Tian, Q.: Higcin: Hierarchical graph-based
  cross inference network for group activity recognition. IEEE transactions on
  pattern analysis and machine intelligence  (2020)

\bibitem{yan2020social}
Yan, R., Xie, L., Tang, J., Shu, X., Tian, Q.: Social adaptive module for
  weakly-supervised group activity recognition. In: European Conference on
  Computer Vision. pp. 208--224. Springer (2020)

\bibitem{yang2019condconv}
Yang, B., Bender, G., Le, Q.V., Ngiam, J.: Condconv: Conditionally
  parameterized convolutions for efficient inference (2019)

\bibitem{ye2020dynamic}
Ye, F., Pu, S., Zhong, Q., Li, C., Xie, D., Tang, H.: Dynamic gcn:
  Context-enriched topology learning for skeleton-based action recognition. In:
  Proceedings of the 28th ACM International Conference on Multimedia. pp.
  55--63 (2020)

\end{thebibliography}
\end{document}